\documentclass{ieeeaccess}

\usepackage{xcolor}
\usepackage{amsthm}
\usepackage{multirow}
\usepackage{booktabs}
\usepackage{subcaption}
\usepackage{enumitem}
\usepackage{url} 
\usepackage{longtable}
\usepackage{caption}
\usepackage{hyperref}
\usepackage{bm}
\usepackage{soul,xcolor} 
\sethlcolor{yellow}      
\usepackage{mathtools}

\usepackage{algorithm}
\usepackage{algpseudocode}
\usepackage{array}
\usepackage{makecell}

\usepackage{cite}
\usepackage{amsmath,amssymb,amsfonts}
\usepackage{graphicx}
\usepackage{textcomp}
\usepackage{float}
\def\BibTeX{{\rm B\kern-.05em{\sc i\kern-.025em b}\kern-.08em
    T\kern-.1667em\lower.7ex\hbox{E}\kern-.125emX}}
\begin{document}
\history{Date of publication xxxx 00, 0000, date of current version xxxx 00, 0000.}
\doi{10.1109/ACCESS.2017.DOI}

\title{HyBattNet: Hybrid Framework for Predicting the Remaining Useful Life of Lithium-Ion Batteries}

\author{
\uppercase{Khoa Tran\authorrefmark{1}}, 
\uppercase{Tri Le\authorrefmark{1}}, 
\uppercase{Bao Huynh\authorrefmark{1}}, 
\uppercase{Hung-Cuong Trinh\authorrefmark{2}}, 
\uppercase{Vy-Rin Nguyen\authorrefmark{3}}, 
\uppercase{T. Nguyen-Thoi\authorrefmark{4}\authorrefmark{5}}, 
\uppercase{Vin Nguyen-Thai\authorrefmark{6}\authorrefmark{7}}
}

\address[1]{AIWARE Limited Company, 17 Huynh Man Dat Street, Hoa Cuong Bac Ward, Hai Chau District, Da Nang 50000, Vietnam (e-mails: khoa.tran@aiware.website; tri.le@aiware.website; bao.huynh@aiware.website)}
\address[2]{Natural Language Processing and Knowledge Discovery Research Group, Faculty of Information Technology, Ton Duc Thang University, Ho Chi Minh City 70000, Vietnam (e-mail: trinhhungcuong@tdtu.edu.vn)}
\address[3]{Software Engineering Department, FPT University, Da Nang 50000, Vietnam (e-mail: RinNV@fe.edu.vn)}
\address[4]{Laboratory for Applied and Industrial Mathematics, Institute for Computational Science and Artificial Intelligence, Van Lang University, Ho Chi Minh City 70000, Viet Nam (e-mail: trung.nguyenthoi@vlu.edu.vn)}
\address[5]{Faculty of Mechanical, Electrical, and Computer Engineering, Van Lang School of Technology, Van Lang University, Ho Chi Minh City 70000, Viet Nam}
\address[6]{Laboratory for Computational Mechanics, Institute for Computational Science and Artificial Intelligence, Van Lang University, Ho Chi Minh City 70000, Viet Nam (e-mail: vin.nguyenthai@vlu.edu.vn)}
\address[7]{Faculty of Civil Engineering, Van Lang School of Technology, Van Lang University, Ho Chi Minh City 70000, Viet Nam}

\tfootnote{Khoa Tran, Tri Le, and Bao Huynh contributed equally to this work.}

\markboth
{Tran \headeretal: An Hybrid Framework for ...}
{Tran \headeretal: An Hybrid Framework for ...}

\corresp{Corresponding author: Hung-Cuong Trinh (e-mail: trinhhungcuong@tdtu.edu.vn).}




\begin{abstract}
Accurate prediction of the Remaining Useful Life (RUL) is essential for enabling timely maintenance of lithium-ion batteries, impacting the operational efficiency of electric applications that rely on them. This paper proposes a RUL prediction approach that leverages data from recent charge-discharge cycles to estimate the number of remaining usable cycles. The approach introduces both a novel signal preprocessing pipeline and a deep learning prediction model. In the signal preprocessing pipeline, a derived capacity feature is computed using interpolated current and capacity signals. Alongside original capacity, voltage and current, these features are denoised and enhanced using statistical metrics and a delta-based method to capture differences between the current and previous cycles. In the prediction model, the processed features are then fed into a hybrid deep learning architecture composed of 1D Convolutional Neural Networks (CNN), Attentional Long Short-Term Memory (A-LSTM), and Ordinary Differential Equation-based LSTM (ODE-LSTM) blocks. The ODE-LSTM architecture employs ordinary differential equations to integrate continuous dynamics into sequence-to-sequence modeling, thereby combining continuous and discrete temporal representations, while the A-LSTM incorporates an attention mechanism to capture local temporal dependencies. The model is further evaluated using transfer learning across different learning strategies and target data partitioning scenarios. Results indicate that the model maintains robust performance, even when fine-tuned on limited target data. Experimental results on two publicly available LFP/graphite lithium-ion battery datasets demonstrate that the proposed method outperforms a baseline deep learning approach and machine learning techniques, achieving an RMSE of 101.59, highlighting its potential for real-world RUL prediction applications.
\end{abstract}


\titlepgskip=-15pt

\maketitle

\section{Introduction}
\PARstart{T}{he} growing global emphasis on sustainable energy has accelerated the advancement of battery technologies in recent years. Consequently, the lithium-ion battery (LIB) market is projected to surpass 170 billion USD by 2030~\cite{meng2019review}. LIBs are widely used in applications such as grid energy storage systems, electric vehicles (EVs), and drones, all of which increasingly depend on effective Battery Health Management (BHM) systems~\cite{sierra2019battery, che2025unlocking, jiao2023lightgbm} to ensure reliable performance. Typical BHM systems include key functionalities such as state of charge (SOC) estimation~\cite{haraz2025ensemble, peng2024state, shah2024novel}, state of health (SOH) estimation~\cite{sun2024state, ma2023estimating, du2025feature}, and remaining useful life (RUL) prediction~\cite{shi2026enhanced, dong2026load, ma2025enhanced, zhao2025rul, zraibi2025online, jia2025joint, alharbi2025lithium}. Among these, RUL prediction—which estimates the number of charge-discharge cycles remaining before a battery reaches its end-of-life (EOL)—is particularly crucial due to its implications for proactive maintenance, and safety assurance. During operation, LIBs undergo multiple forms of degradation during cycling, primarily caused by the loss of lithium inventory (LLI) and loss of active material (LAM)~\cite{ma2024accurate}. These challenges underscore the need for accurate RUL prediction models, not only to support real-time diagnostics and timely maintenance, but also to inform battery design~\cite{ji2023deep}. In modern RUL prediction for LIBs, methods are mainly classified into two categories: curve-based and cycle-feature-based approaches. 

Curve-based methods model the degradation of a health indicator (e.g., capacity) over cycles using predefined functions, such as exponential decay models like the double exponential model (DEM)~\cite{ma2023two}, variational mode decomposition (VMD)~\cite{gao2024multi}, or empirical mode decomposition (EMD)~\cite{liu2024hybrid, li2023development}. \cite{deng2023prognostics} proposed a data-driven capacity prognostic method for on-road EV batteries using only charging data, combining ampere-hour integral capacity estimation, statistical denoising of monthly labels, and a Seq2Seq model with Gaussian Process residuals. While their approach achieved less than 1.6\% prediction error across 29 months of real-world EV data, the framework primarily focused on charging-derived features and did not fully account for discharge dynamics or cell-to-cell variability. \cite{wu2025ldnet} introduced LDNet-RUL, a lightweight deformable neural network designed for capacity prediction of lithium-ion batteries.  
By incorporating deformable convolution modules to adaptively capture local degradation patterns, their model achieves competitive accuracy with reduced computational cost, though its reliance on handcrafted input features and relatively small benchmark datasets may limit scalability to diverse real-world operating conditions.  By fitting the historical data, these methods extrapolate the cycle at which the indicator falls below a predefined EOL threshold (e.g., 80\% capacity~\cite{severson2019data}), thereby yielding the predicted number of remaining charge-discharge cycles (RUL). However, these methods have limited flexibility, as they assume a fixed degradation trend throughout the entire battery cycle life, which may not accurately capture real-world variability. Moreover, they often require a long degradation history to make accurate predictions. For instance, the method in~\cite{wang2023remaining} relies on data from the first 450 charge-discharge cycles to perform accurate RUL prediction, making it unsuitable for early-stage prognostics. In contrast, cycle-feature-based methods extract meaningful features from a few recent cycles~\cite{xia2023historical, ma2022real}, enabling more adaptive, earlier predictions under different operating conditions across charge-discharge cycles.

In recent cycle-feature-based approaches, \cite{xia2023historical} predicts RUL of LIBs using only features extracted from the current charge-discharge cycle. The features include discharge capacity, initial battery capacity, and working condition-derived variables. However, this approach neglects the short-term degradation trends observable in the preceding cycles, which are critical for accurately capturing the battery’s dynamic aging behavior. Although \cite{xia2023historical} reports promising results, these outcomes are partially influenced by an imbalanced data split—70 batteries for training and only 7 for testing—which makes the testing set small compared to the training set, resembling a biased setup that may lead to overfitting and reduced generalizability. In another study, \cite{zhang2019synchronous} employs features extracted from the incremental capacity (IC) curve of a partial segment of the current charge-discharge cycle to predict the SOH and RUL. However, the method is evaluated only on a small battery dataset comprising just four battery cells discharged under constant current. This results in highly similar data distributions between the training and testing sets, making the prediction task easier and less representative of real-world scenarios. In contrast, a more robust and realistic approach is presented in \cite{ma2022real}, where features are extracted from 10 selected cycles within the 30 most recent charge-discharge cycles to predict RUL. This strategy effectively captures recent degradation patterns, enhancing predictive performance. Furthermore, \cite{ma2022real} incorporates transfer learning to leverage degradation knowledge from a different source domain, improving the prediction accuracy in the target domain. Their method is validated on one of the largest publicly available battery datasets, comprising 55 cells for training and 22 cells for testing, with diverse discharge profiles to simulate real-world conditions.

Taken together, these related works highlight several important limitations of existing cycle-feature-based RUL prediction methods. First, approaches either rely on features from only a single current cycle~\cite{xia2023historical, zhang2019synchronous} or operate under highly constrained operating conditions and small cell populations, which limits their generalizability. Second, although~\cite{ma2022real} represents a significant step forward by leveraging features from multiple recent cycles and incorporating transfer learning on a larger benchmark, its feature engineering and model structure still rely on relatively shallow temporal modeling. These gaps motivate the need for a hybrid framework that (i) has a signal preprocessing pipeline that exploits rich short-term degradation information from a window of recent cycles and feeds it into a more expressive temporal deep learning prediction model, and (ii) is validated on large-scale datasets under realistic data splits to demonstrate robust generalization.

In this work, we adopt the standard approach to data splitting, input representation, and target definition as introduced in the aforementioned study~\cite{ma2022real}, where a window of recent sequential charge-discharge cycles is used as input to predict the number of remaining charge-discharge cycles. Our proposed method features a novel signal preprocessing pipeline and a hybrid deep learning (DL) prediction model. 

The signal preprocessing pipeline consists of four main steps: Capacity Interpolation and Denoising, Statistical Feature Extraction, Delta Feature Computation, and Feature Fusion. These steps extract a tensor input sample that encapsulates informative features from a recent window of charge-discharge cycles.

The prediction model comprises two parallel branches. The first branch integrates a 1D Convolutional Neural Network (CNN) block and Attentional Long Short-Term Memory (A-LSTM) network~\cite{liu2020anomaly} to capture temporal dependencies. The second branch employs an Ordinary Differential Equation-based LSTM (ODE-LSTM) block to enhance the modeling of dynamic sequences. The output from both branches are fused and passed through a fully connected layer to produce a RUL value, representing the predicted number of remaining charge-discharge cycles. 

We also evaluate the proposed prediction model using transfer learning under various learning strategies and target data splitting scenarios. The results demonstrate that our model remains robust even when trained on limited target data. The model is validated on two of the most widely used lithium-ion battery datasets: one with varying charging profiles~\cite{severson2019data} and the other with diverse discharging profiles~\cite{ma2022real}.

The remainder of this paper is organized as follows. Section~\ref{sec:proposed_method} introduces the proposed method, including signal preprocessing, model architecture, and evaluation metrics. Section~\ref{sec:dataset} describes the datasets and presents data analysis. Section~\ref{sec:experiments} covers the experimental setup, model evaluations, additional validation with transfer learning, and discussion. Finally, Section~\ref{sec:conclusion} concludes the paper and outlines future work.

\section{Proposed Method}\label{sec:proposed_method}
The proposed method predicts RUL of lithium-ion batteries, defined as the number of charge-discharge cycles remaining from the current cycle $i$ until the battery reaches its EOL. The approach comprises two main stages: (1) a signal preprocessing approach, and (2) a prediction model. 


We implement the entire pipeline in Python. Signal preprocessing uses NumPy~\cite{harris2020array}, SciPy~\cite{virtanen2020scipy}, pandas~\cite{reback2020pandas}, and scikit-learn~\cite{pedregosa2011scikit}. Feature fusion and batching are handled with NumPy and the PyTorch DataLoader~\cite{imambi2021pytorch}. The prediction model is built and trained in PyTorch, with ODE solvers from torchdiffeq~\cite{kidger2021hey}; training runs on CUDA/cuDNN-enabled GPUs, and plots are produced with Matplotlib~\cite{bisong2019matplotlib}.

\begin{figure*}[!t]
  \centering
  \includegraphics[width=1\textwidth]{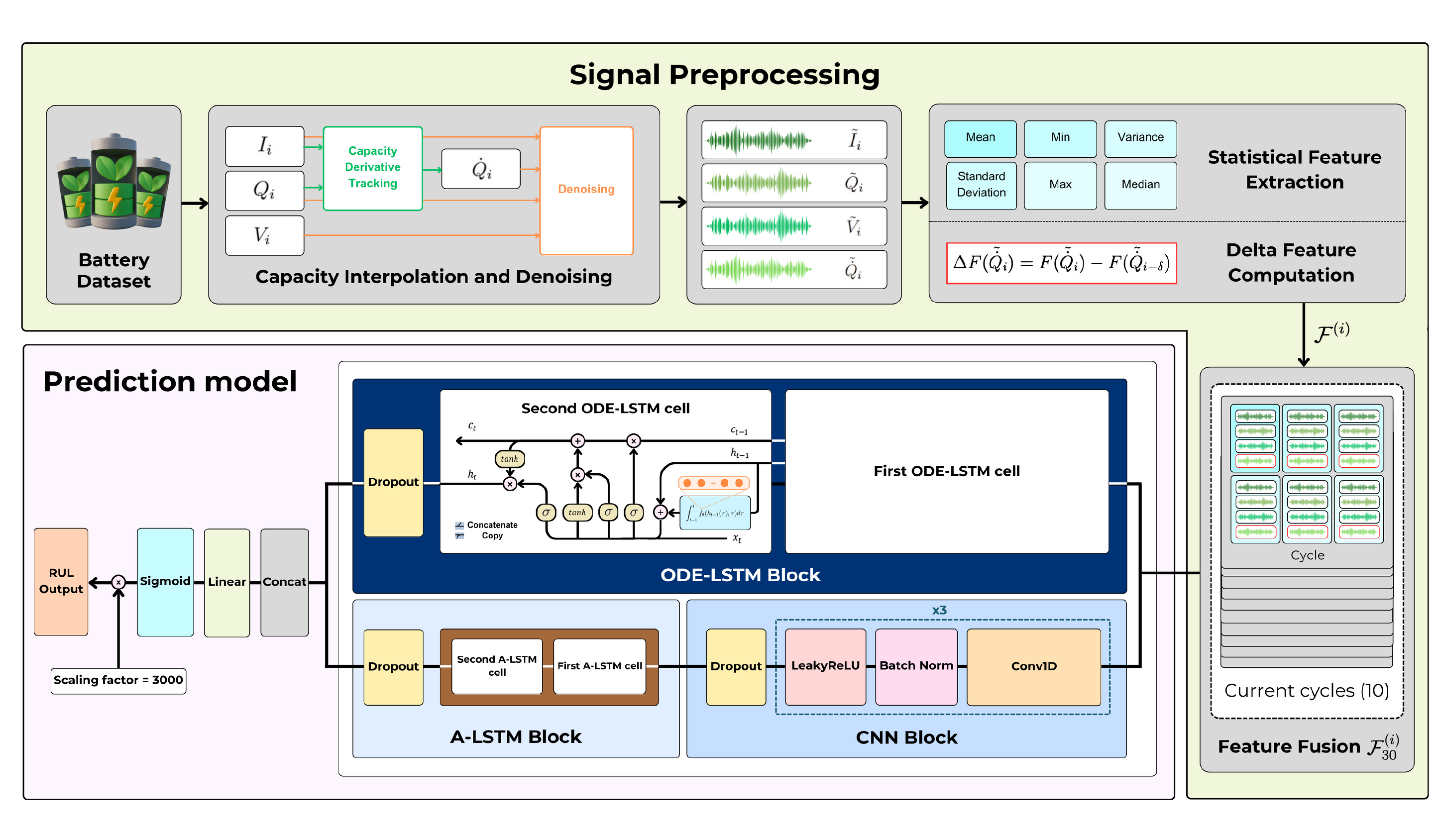}
  \caption{Overall architecture of our proposed hybrid framework.}
  \label{fig:Main_architecture}
\end{figure*}

The overall architecture is illustrated in Figure~\ref{fig:Main_architecture}. The details of components are presented in the following subsections.

\subsection{Signal Preprocessing}
We propose a signal preprocessing approach that, for each charge–discharge cycle \(i\), constructs a input sample \(\mathcal{F}^{(i)}_{30}\) for the neural network. The pipeline aims to denoise the raw signals and emphasize salient characteristics, thereby improving computational efficiency and predictive performance. The proposed signal preprocessing pipeline consists of the following four phases.

\paragraph{Capacity Interpolation and Denoising} 
For each charge--discharge cycle \(i\), the raw current \(I_i \in \mathbb{R}^{\ell}\), voltage \(V_i \in \mathbb{R}^{\ell}\), and capacity \(Q_i \in \mathbb{R}^{\ell}\) are first processed by the Capacity Interpolation and Denoising phase. This phase computes a derived capacity \(\dot{Q}_i \in \mathbb{R}^{n}\) and subsequently outputs denoised sequences. Here, \(\ell\) denotes the original sequence length. The detailed procedure is as follows.

Given the original current--capacity samples \(\{(I_{i,j},\, Q_{i,j})\}_{j=1}^{\ell}\), sorted in ascending order of \(I_{i,j}\), we apply one-dimensional linear interpolation~\cite{numpy_interp} on a uniform current grid \(\mathcal{I}\) to obtain the derived capacity (hereafter referred to as \emph{Capacity Derivative Tracking}):
\[
\dot{Q}_i
= \operatorname{Interp1D}\!\big(I_{i,j},\, Q_{i,j}\big)\big|_{I_k},
\qquad \forall\, I_k \in \mathcal{I},
\]
where \(\operatorname{Interp1D}(\cdot)\) denotes the  one-dimensional
interpolation operator (a spline in our case). The grid \(\mathcal{I}\) is
defined as a set of evenly spaced current values:
\[
\mathcal{I}
= \big\{\, I_k \,\big|\, I_k \in [I_i^{\min},\, I_i^{\max}],\; k = 1,\dots,n \big\},
\]
where \(I_i^{\min}\) and \(I_i^{\max}\) are the minimum and maximum of the
current signal \(I_i\), respectively, and \(n\) is the number of interpolation points (we use $n=2000$ after grid search over \(\{1000, 1500, 2000, 2057\}\); here, \(2057\) is the minimum length of the raw sequences in the training data). Across cycles, the lengths of the raw capacity sequences vary due to battery aging, usage deviations, and operating conditions. The \emph{Capacity Derivative Tracking} method (which produces \(\dot{Q}_i\)) standardizes these variable-length sequences to a common length \(n\), mapping heterogeneous capacity sequences into a shared representation space. This normalization facilitates the detection of aging patterns and improves the model’s RUL prediction accuracy. The effectiveness of \(\dot{Q}_i\) is demonstrated in Subsection~\ref{subsec:prediction}.

The resulting sequences $I_i$, $V_i$, $Q_i$, and $\dot{Q}_i$ are then denoised using a Savitzky–Golay filter~\cite{schafer2011savitzky}, yielding the smoothed sequences:
\[
\tilde{x}_i \in \left\{ \tilde{I}_i, \tilde{V}_i, \tilde{Q}_i, \tilde{\dot{Q}}_i \right\}.
\]

\paragraph{Statistical Feature Extraction}\label{subsec:feature_extraction}
In this phase, each denoised sequence \(\tilde{x}_i \in \mathbb{R}^{T}\) is processed by a feature-extraction operator \(F(\cdot)\) to compute a six-dimensional vector of statistical descriptors.
Formally, let \(F:\mathbb{R}^{T}\!\to\mathbb{R}^{6}\) and define
\begin{align}
F(\tilde{x}_i) = \big[ & \mu(\tilde{x}_i),\ \sigma(\tilde{x}_i),\ \min(\tilde{x}_i), \notag \\
                       & \max(\tilde{x}_i),\ \mathrm{Var}(\tilde{x}_i),\ \mathrm{Median}(\tilde{x}_i) \big],
\end{align}
where $T$ is the sequence length of the denoised vector $\tilde{x}_i$, $\mu$ is the mean, $\sigma$ is the standard deviation, $\min$ and $\max$ denote the minimum and maximum values, $\mathrm{Var}$ is the variance, and $\mathrm{Median}$ is the median of the smoothed sequence $\tilde{x}_i$. These descriptors reduce computational complexity by compressing each sequence (\(>2000 \) data points) into six scalars, while retaining the information needed for strong predictive performance.

\paragraph{Delta Feature Computation.}
To capture temporal dynamics across cycles, a lag-\(\delta\) feature difference is defined to quantify inter-cycle change:
\begin{equation}
\Delta F_i \coloneqq F(\tilde{\dot{Q}}_i) - F(\tilde{\dot{Q}}_{i-\delta}) \in \mathbb{R}^{6}, \qquad i > \delta,
\end{equation}
where \(\delta = 9\) charge--discharge cycles (chosen based on the best performance observed in the experiments, as shown in Figure~\ref{fig:rmse_combined_delta}) and \(F(\cdot) \in \mathbb{R}^{6}\) is the feature-extraction operator defined in Subsection~\ref{subsec:feature_extraction}.
For \(i \le \delta\), \(\Delta F_i\) is omitted to maintain consistent tensor shapes. 

This delta-feature computation has also been employed in prior RUL prediction studies in aerospace systems~\cite{peng2022remaining, ensariouglu2023remaining}, to capture temporal evolution while balancing sensitivity to short-term fluctuations and long-term degradation trends. 

In our study, we performed a grid search over delta feature computation applied to each feature \(\dot{Q}, Q, I, V\) and their combinations. The best validation performance (RMSE) was obtained when applying the delta feature computation only to \(\dot{Q}\), indicating that this choice preserves the most predictive signal for RUL.



\paragraph{Feature Fusion.}
The features
\(F(\tilde{I}_i)\), \(F(\tilde{V}_i)\), \(F(\tilde{Q}_i)\), and \(\Delta F_i\), collected in the previous phases, are stacked row-wise to yield the fused feature matrix
\[
\mathcal{F}^{(i)} \coloneqq
\begin{bmatrix}
F(\tilde{I}_i)\\
F(\tilde{V}_i)\\
F(\tilde{Q}_i)\\
\Delta F_i
\end{bmatrix}
\in \mathbb{R}^{4\times 6}.
\]
The fused feature matrices \(\mathcal{F}^{(\cdot)}\) from 10 uniformly spaced cycles within a 30-cycle window (i.e., one every three cycles) are combined to form a input sample, following the strategy in~\cite{ma2022real}. The resulting input tensor is
\[
    \mathcal{F}^{(i)}_{30} \in \mathbb{R}^{10 \times 4 \times 6}.
\]
$\mathcal{F}^{(i)}_{30}$ is then reshaped to $\mathbb{R}^{10 \times 24}$ for MinMax scaling~\cite{patro2015normalization} to the range $(0, 1)$ across the 24 dimensions, then reshaped back to $\mathbb{R}^{10 \times 4 \times 6}$. The scaler is fitted on the training set and applied unchanged to validation/test. 

Note that for a input sample anchored at cycle \(i\), all features are computed using only cycles in \([i-29,\, i]\). We construct input samples only when \(i-29 > \delta\).

\subsection{Prediction Model}
Each input sample $\mathcal{F}^{(i)}_{30}$, as described in the previous section, is fed into the proposed prediction model to estimate the RUL. Our prediction model comprises three primary components: CNN block, A-LSTM block, and ODE-LSTM block. Among these, A-LSTM block is adopted from the prior work~\cite{liu2020anomaly}, while the other components are tailored.

\paragraph{CNN block} 
The input sample $\mathcal{F}^{(i)}_{30}\in\mathbb{R}^{10\times4\times6}$ is reshaped into a matrix of shape $\mathbb{R}^{10\times24}$ by flattening the $4\times6$ features at each time step. This matrix is then passed through a stack of three one-dimensional convolutional layers~\cite{kiranyaz20211d} with a kernel size of 5 and progressively increasing channel dimensions—$H$, $2H$, and $4H$, where $H = 64$—consistent with the optimal configuration identified in Figure~\ref{fig:rmse_model_configs}. Each convolution operation is followed by batch normalization~\cite{bjorck2018understanding}, and a Leaky ReLU activation~\cite{xu2020reluplex}. A dropout~\cite{wei2020implicit} layer with a fixed rate of 0.3 is applied at the end of the CNN block for regularization. Note that this fixed dropout rate of 0.3 is consistently used across all dropout layers in the prediction model. The resulting output is
\[
X_{\mathrm{CNN}} = \mathrm{CNN}\!\big(\mathcal{F}^{(i)}_{30}\big) \in \mathbb{R}^{10 \times 256},
\]
which captures temporal dependencies and local patterns within the cycle window. This output is then passed to the A-LSTM block.


\paragraph{A-LSTM Block}
The CNN output \(X_{\mathrm{CNN}}\in \mathbb{R}^{10\times 256}\) is fed into two A-LSTM layers~\cite{liu2020anomaly} (hidden size 128) stacked and applied over the time sequence. At time step \(t\), let \(x'_t \coloneqq X_{\mathrm{CNN}}[t,:]\in \mathbb{R}^{256}\). The recurrences are
\begin{align}
(h_t^{(1)}, c_t^{(1)}) 
&= \operatorname{A\mbox{-}LSTM}^{(1)}\!\left(
      x'_t,\, h_{t-1}^{(1)},\, c_{t-1}^{(1)};\, T_t
   \right), \\
(h_t^{(2)}, c_t^{(2)}) 
&= \operatorname{A\mbox{-}LSTM}^{(2)}\!\left(
      h_t^{(1)},\, h_{t-1}^{(2)},\, c_{t-1}^{(2)};\, T_t
   \right), \\
\mathbf{z} &= \operatorname{Dropout}_{0.3}\!\left(h_T^{(2)}\right).
\end{align}
\noindent
With sequence length \(T=10\), the hidden and cell states satisfy
\(h_t^{(m)},\, c_t^{(m)} \in \mathbb{R}^{128}\) for \(t=1,\dots,10\) and \(m\in\{1,2\}\), where \(m\) indexes the layer. We initialize \(h_0^{(m)}=\mathbf{0}\) and \(c_0^{(m)}=\mathbf{0}\). The trend-attention gate (TAG) \(T_t\) is computed from \(x'_t\) and its local waveform \(R_t\) (defined above) as
\[
T_t=\sigma\!\big(W_x x'_t + \sigma(W_r R_t) + b_t\big),
\]
and it modulates the gates within each A-LSTM cell. Unlike a standard LSTM, where $o_t$ and $c_t$ depend only on the input, hidden state, and learned gates, the A-LSTM modulates them with $T_t$, enabling the model to emphasize more informative timesteps and attenuate less relevant ones. The final output of the A-LSTM block is
\[
X_{\text{A-LSTM}}
= \mathbf{z}
\in \mathbb{R}^{128}.
\]

\paragraph{ODE-LSTM Block}
The input sample \(\mathcal{F}^{(i)}_{30}\in \mathbb{R}^{10\times 24}\) is fed into two stacked ODE-LSTM layers (input size \(24\), hidden size \(256\)) and processed over a time sequence of length \(T=10\). At time step \(t\), let
\(f'_t \coloneqq \mathcal{F}^{(i)}_{30}[t,:]\in \mathbb{R}^{24}\).
The recurrences are
\begin{align}
\tilde{h}_{t} &= h_{t-1} + \operatorname{ODE}_{\theta}\!\left(h_{t-1};\, t-1,\, t\right), \\
i_{t} &= \sigma\!\big(W_i [f'_t;\tilde{h}_t] + b_i\big), \\
f_{t} &= \sigma\!\big(W_f [f'_t;\tilde{h}_t] + b_f\big), \\
o_{t} &= \sigma\!\big(W_o [f'_t;\tilde{h}_t] + b_o\big), \\
\tilde{c}_{t} &= \tanh\!\big(W_c [f'_t;\tilde{h}_t] + b_c\big), \\
c_{t} &= f_t \odot c_{t-1} + i_t \odot \tilde{c}_t, \\
h_{t} &= o_t \odot \tanh(c_t), \\
\tilde{\mathbf{z}} &= \operatorname{Dropout}_{0.3}\!\left(h_T\right).
\end{align}
\noindent
The hidden and cell states satisfy
\(h_t,\, c_t \in \mathbb{R}^{256}\) for \(t=1,\dots,10\).
We initialize \(h_0=\mathbf{0}\) and \(c_0=\mathbf{0}\).
The continuous transition is
\[
\operatorname{ODE}_{\theta}(h;\, t_0, t_1)
:= \int_{t_0}^{t_1} f_{\theta}(h,t)\,dt
\;\approx\; (t_1-t_0)\, f_{\theta}(h,t_0),
\]
where \(f_{\theta}\) is a neural network with hidden size \(256\). This formulation allows the latent dynamics of the network to evolve smoothly
between two adjacent sampling points. In the context of battery degradation, the internal electrochemical states do not change abruptly but evolve gradually, governed by continuous physical processes such as ion diffusion, and solid–electrolyte interphase (SEI) growth~\cite{wang2018review}. This enables the continuous transition model to capture variable-rate degradation behaviors and preserve the temporal continuity of health indicators. \([\,\cdot;\cdot\,]\) denotes vector concatenation, and \(\odot\) denotes element-wise multiplication. The final output of the ODE-LSTM block is
\[
X_{\text{ODE-LSTM}} \;=\; \tilde{\mathbf{z}} \in \mathbb{R}^{256}.
\]

The output features from the CNN\,+\,A-LSTM and ODE--LSTM branches are concatenated into a single vector of dimension \(384\), which is then passed through a linear layer followed by a sigmoid activation:
\[
\begin{aligned}
\mathbf{u} &= [X_{\text{A-LSTM}};\, X_{\text{ODE-LSTM}}] \in \mathbb{R}^{384}, \\
\hat{y} &= 3000 \cdot \sigma(W\mathbf{u} + b).
\end{aligned}
\]
Here, the factor \(3000\) scales the sigmoid output from \([0,1]\) to \([0,3000]\) \emph{cycles}, matching the maximum expected RUL range; thus \(\hat{y}\) represents the predicted number of useful remaining cycles (RUL).

Our proposed hybrid deep learning model is trained using Mean Squared Error (MSE) loss to remain consistent with most previous approaches and ensure a fair comparison. In future work, we plan to investigate heteroscedastic loss formulations. The model is optimized with the AdamW optimizer~\cite{loshchilov2017decoupled}. The MSE loss is defined as:
\[
\mathcal{L}_{\text{MSE}} = \frac{1}{N} \sum_{i=1}^{N} \left( y_i - \hat{y}_i \right)^2.
\]
Where $N$ is the number of training samples, $y_i$ denotes the ground truth RUL of the $i$-th cycling sample, and $\hat{y}_i$ denotes the predicted RUL of the $i$-th sample.

\subsection{Evaluation Metrics}
To assess the performance of RUL prediction, we employ three evaluation metrics: Root Mean Square Error (\(RMSE\)) \cite{willmott2005advantages}, R-squared (\(R^2\)) \cite{wang2023remaining}, and Mean Absolute Percentage Error (\(MAPE\)) \cite{de2016mean}. These metrics are defined as follows:

\[
RMSE(y_i, \hat{y}_i) = \sqrt{\frac{1}{n} \sum_{i=1}^{n} (y_i - \hat{y}_i)^2},
\]

\[
MAPE(y_i, \hat{y}_i) = \frac{1}{n} \sum_{i=1}^{n} \frac{|y_i - \hat{y}_i|}{y} \times 100,
\]

\[
R^2(y, \hat{y}) = 1 - \frac{\sum_{i=1}^{n} (y_i - \hat{y}_i)^2}{\sum_{i=1}^{n} (y_i - \bar{y})^2}.
\]

Here, \( y_i \) represents the observed RUL, and \( \hat{y}_i \) represents the predicted RUL. Additionally, \(y\) refers to the cycle life. The performance improves with lower values of \(RMSE\) and \(MAPE\), and a higher value of \(R^2\).

\section{Dataset}\label{sec:dataset}
\subsection{Dataset Description}
\paragraph{First dataset}
The dataset~\cite{severson2019data} features a comprehensive study of 124 lithium-ion batteries (LIBs) with LFP/graphite chemistry, focusing on various fast-charging conditions while maintaining consistent discharging parameters. Each battery in the dataset has a nominal capacity of 1.1 Ah and a nominal voltage of 3.3 V, with cycle lifespans ranging from 150 to 2,300 cycles, highlighting a broad range of battery longevity. All LIBs underwent uniform discharge procedures, involving a constant current discharge at a rate of 4 C until the voltage reached 2 V, followed by a constant voltage discharge at 2 V until the current dropped to C/50. Charging was conducted at rates between 3.6 C and 6 C under controlled temperature conditions of 30°C in an environmental chamber. The dataset comprises roughly 96,700 cycles, making it one of the largest datasets exploring various fast-charging protocols. 

In this dataset, battery EOL is defined at 80\% of the nominal capacity (i.e., 0.88 Ah for these cells). While the dataset provides extensive coverage across fast-charging policies, it is imbalanced with respect to cycle life: many cells reach EOL relatively early (150–600 cycles), whereas only a smaller subset survives beyond 2,000 cycles. This imbalance poses challenges for model generalization, as predictive accuracy can be biased toward short-lived cells.

Following the official split reported in the Supplementary Information of \cite{severson2019data}, we use 41 LIBs for training, 43 for validation, and 40 for testing. The training and validation sets are drawn from the same manufacturing batches (2017-05-12 and 2017-06-30) and the same charging-policy families, but are disjoint by cell. For most policies, one LIB is assigned to the training set and its matched counterpart to the validation set, yielding an in-distribution holdout of unseen cells. The test set comprises all LIBs from the later 2018-04-12 batch, serving as an out-of-distribution holdout due to batch and protocol shift.

\begin{figure*}[!t]
  \centering
  \includegraphics[width=1\textwidth]{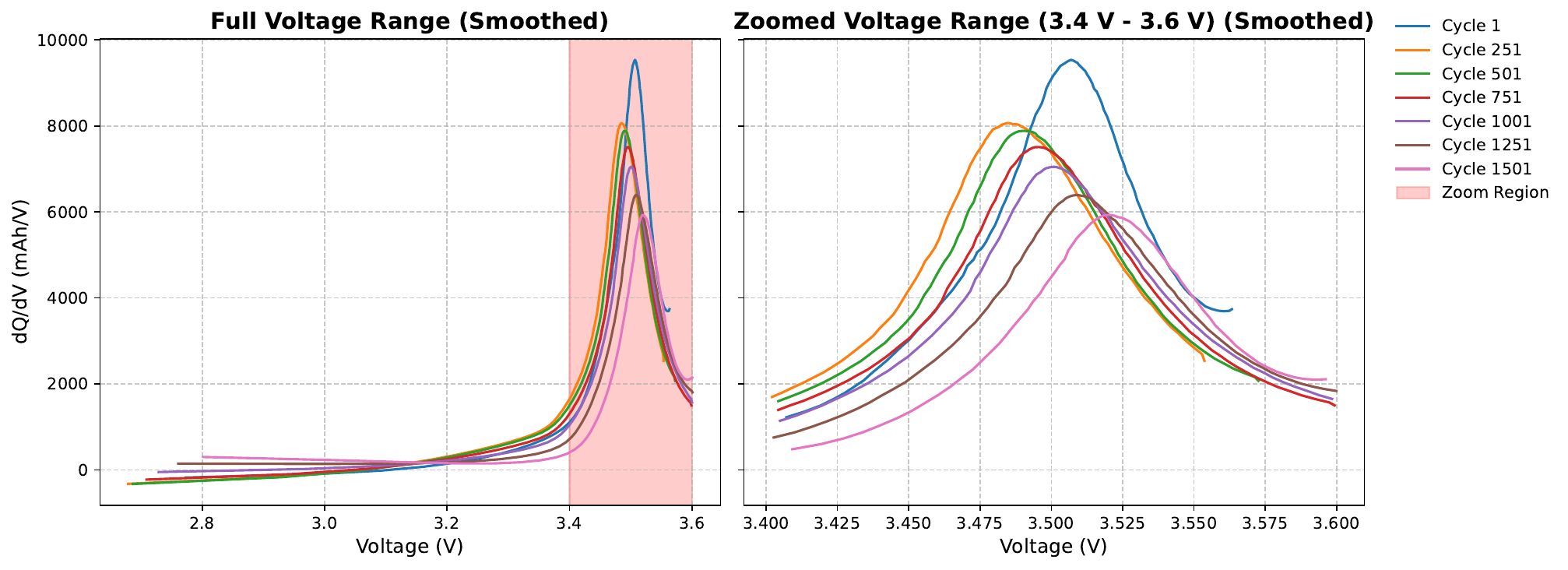}
  \caption{Smoothed incremental capacity (dQ/dV) curves during charging for selected cycles on a battery cell from the second dataset: full voltage range (left) and zoomed-in region (3.4–3.6 V, right).}
  \label{fig:incremental_capacity_degradation}
\end{figure*}

\paragraph{Second dataset}
The dataset~\cite{ma2022real}, which involved varying discharge conditions while maintaining consistent fast-charging conditions, was collected from a battery degradation experiment. This experiment utilized 77 cells (LFP/graphite A123 APR18650M1A), each with a nominal capacity of 1.1 Ah and a nominal voltage of 3.3 V. Each cell underwent a distinct multi-stage discharge protocol, while all cells shared the same fast-charging protocol. The experiment was carried out in two thermostatic chambers, with a controlled temperature of 30°C. The dataset contains a total of 146,122 discharge cycles, making it one of the largest datasets that incorporate various discharge protocols. The cycle life of the cells ranges from 1,100 to 2,700 cycles, with an average of 1,898 cycles and a standard deviation of 387 cycles. The discharge capacity as a function of cycle number shows a broad distribution of cycle lives. 

In this dataset, EOL is also defined at 80\% of the nominal capacity (0.88 Ah). Compared to the first dataset, the discharge-protocol variability introduces additional heterogeneity in cycle lengths and degradation trajectories. While many cells operate stably over long periods (surviving more than 2{,}500 cycles), others reach EOL earlier due to aggressive discharge protocols. This imbalance in cycle life distributions challenges the model to learn from both long-lived and short-lived cells.

Following the official split in \cite{ma2022real}, the dataset is divided into two parts: In 55 LIBs predefined for training, 90\% is used for the training set, 10\% for the validation set, and the remaining 22 LIBs are used as the test set. This protocol-disjoint split intentionally induces a distribution shift between training and testing: because each set contains different discharge protocols, their current profiles, voltage–capacity trajectories, cycle lengths, and operating patterns differ. It provides a stricter evaluation of robustness and transferability to unseen operating conditions.

In both datasets, careful control of operating conditions ensures that degradation patterns is attributed primarily to the designed experimental protocols rather than uncontrolled external factors. In the first dataset, charging rate and temperature were tightly regulated, resulting in 96{,}700 cycles and making it one of the largest publicly available resources for studying the interplay between fast-charging strategies and battery aging. In the second dataset, discharge rate and temperature were carefully controlled, yielding 146{,}122 cycles and providing one of the largest resources for investigating the interplay between discharge strategies and battery aging.

\subsection{Data Analysis}

The degradation of LIBs through charge-discharge cycles is primarily caused by the LLI, leading to irreversible capacity fade. IC analysis, which plots $dQ/dV$ versus voltage, is a widely used diagnostic tool for identifying LLI-related degradation~\cite{ansean2019lithium}. 

As shown in Figure~\ref{fig:incremental_capacity_degradation}, smoothed IC curves during constant current charging illustrate degradation trends in lithium-ion batteries. The position, intensity, and shape of $dQ/dV$ peaks provide key insights into electrochemical aging. Notably, the graphite peak associated with the $\mathrm{LiC_6 \rightarrow LiC_{12}}$ phase transition~\cite{Kinetically} shifts left by over 10mV between Cycle 1 and Cycle 1501 (right plot), indicating lithium loss due to solid electrolyte interphase (SEI) formation. This irreversible process consumes lithium ions. Concurrently, the peak height drops by about 35\%, signaling a decline in active lithium available for intercalation~\cite{Pinson_2013,7488267}.

Overall, LLI accounts for approximately 70\% of the total observed capacity fade, based on the analysis of smoothed IC curves in Figure~\ref{fig:incremental_capacity_degradation} using the methodology from~\cite{Pinson_2013}. Additionally, the figure reveals a capacity reduction of around 20\% after 1500 cycles, with the degradation rate varying across the charge-discharge lifespan. These findings highlight the need for a robust RUL prediction method capable of capturing non-uniform aging patterns to reliably predict the RUL of lithium-ion batteries.

\section{Experiments and Discussion}\label{sec:experiments}
\subsection{Experimental Setup}
Our proposed RUL model is developed using the PyTorch framework. All experiments are performed on an NVIDIA 4090 GPU with 24GB of memory. Each experiment is trained for 10 epochs with a batch size of 128 and a constant learning rate of 0.0005. We train the model for 10 epochs, as the compact feature representation produced by our signal preprocessing pipeline allows the network to converge rapidly. Each input signal (e.g., current, voltage) is compressed into six representative points, rather than 100 resamples~\cite{ma2022real} or dozens of handcrafted features~\cite{severson2019data}, which significantly reduces the model’s computational complexity while preserving essential degradation information. We also tested learning rates of 0.001, 0.0005, 0.0002, and 0.0001. 0.0005 consistently provided the best validation and test performance with the most stable training. To minimize variability in the training process, each experiment is repeated 10 times, and the final prediction is calculated as the average of these multiple runs. 

\begin{figure*}[!t]
  \centering
  \includegraphics[width=1\textwidth]{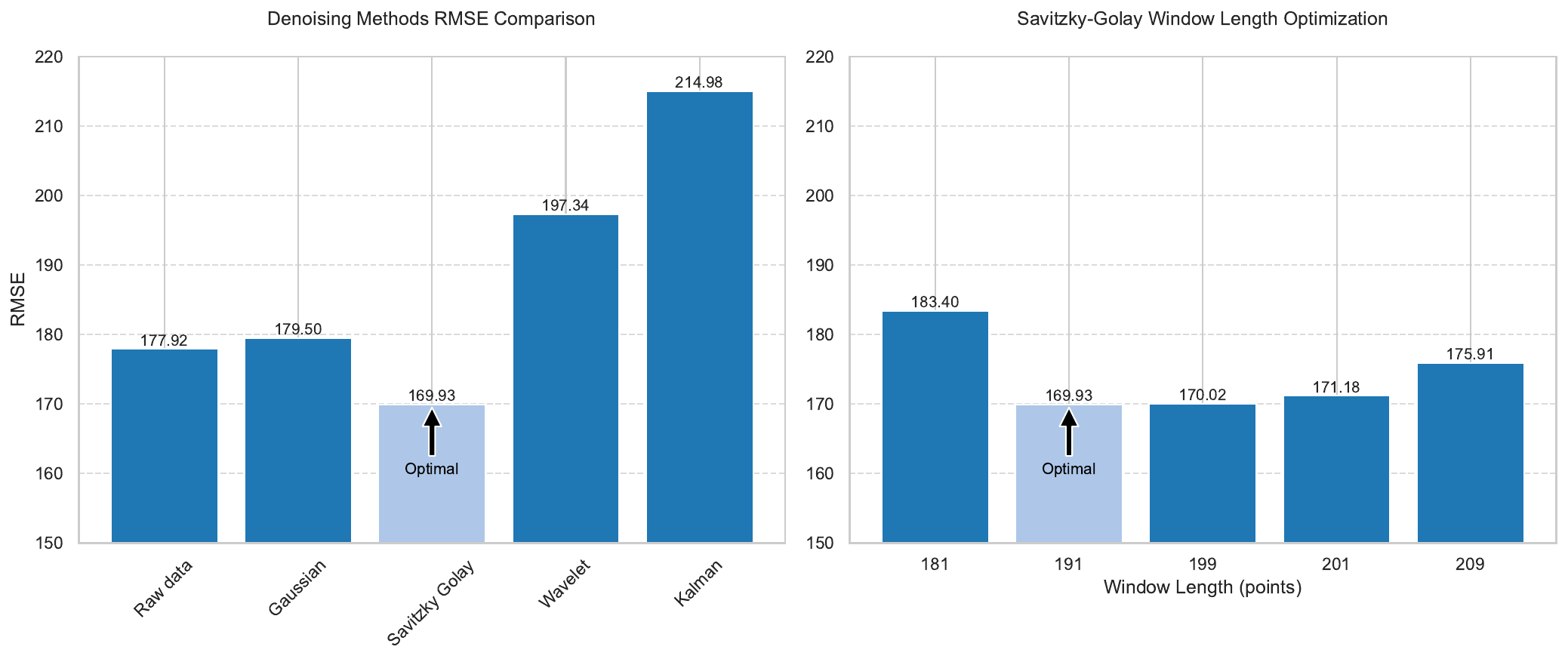}
  \caption{RMSE comparison of different denoising methods (left) and the impact of the window length parameter on Savitzky-Golay method (right).}
  \label{fig:rmse_denoising_comparison}
\end{figure*}

\begin{figure*}[!t]
  \centering
  \includegraphics[width=1\textwidth]{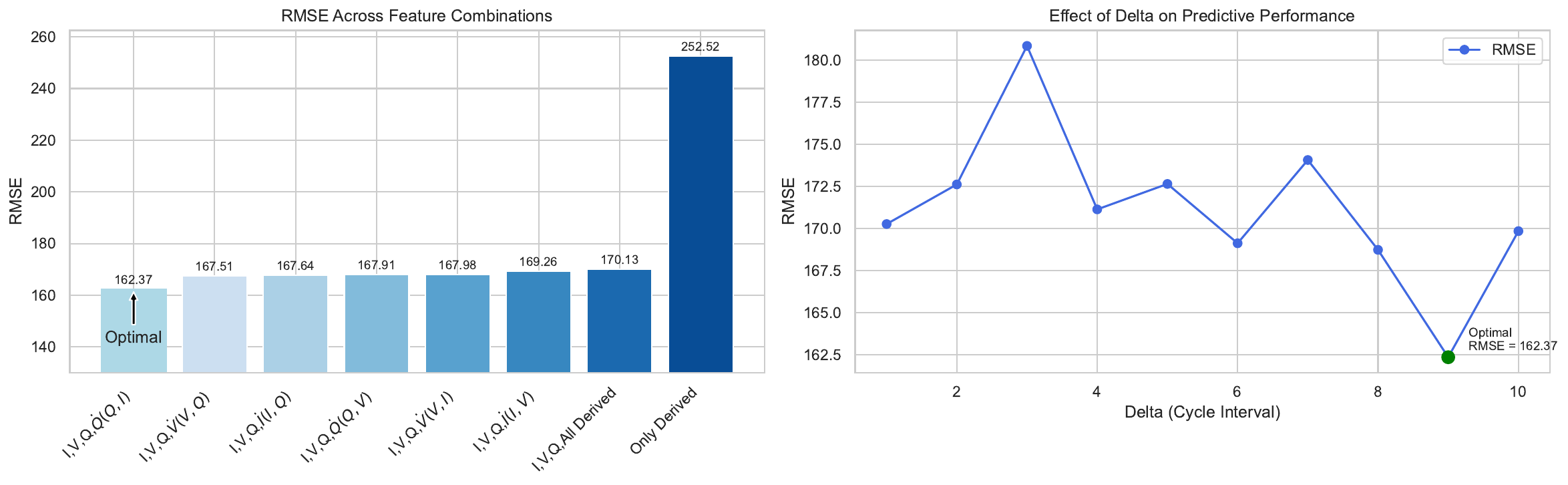}
  \caption{RMSE comparison of feature combinations (left) with \textit{delta} set to 9, and the effect of \textit{delta} on predictive performance (right).}
  \label{fig:rmse_combined_delta}
\end{figure*}

\subsection{Evaluation of Signal Preprocessing}\label{subsec:signal}
In the denoising step, to mitigate noise in voltage, current and capacity signals, four denoising techniques including Gaussian smoothing~\cite{wink2004denoising}, Savitzky-Golay filtering~\cite{karaim2019low}, wavelet decomposition~\cite{ergen2012signal}, and Kalman filtering-along with a no-denoising baseline (raw data), were evaluated on the second dataset. As shown in Figure~\ref{fig:rmse_denoising_comparison}, the Savitzky-Golay filter~\cite{sayadi2008ecg} achieved the optimal performance, yielding an RMSE of 169.93 (cycles). Compared to Gaussian smoothing and wavelet decomposition, the Savitzky-Golay method demonstrates superior capability in peak preservation and adaptive smoothing. 

One key factor influencing the effectiveness of the Savitzky–Golay filter is the window length, which governs the balance between noise reduction and signal distortion. To determine the optimal configuration, various window lengths were evaluated. As shown in the right panel of Figure~\ref{fig:rmse_denoising_comparison}, a window length of 191 points yields the lowest RMSE value, whereas both shorter (181) and longer (199) windows resulted in higher errors. A smaller window length insufficiently smooths the raw signals, leaving residual measurement noise that propagates into feature extraction and model training. Conversely, a larger window over-smooths the signal, attenuating transient behaviors such as end-of-discharge voltage drops. The chosen window length of 191 thus achieves an optimal balance between noise suppression and feature fidelity, preserving the physical integrity that is essential for accurate battery RUL prediction. Consequently, the Savitzky–Golay filter with a window length of 191 is adopted in the denoising stage of our signal preprocessing pipeline.

\begin{table*}[!t]
\centering
\caption{Performance comparison of deep learning models on the testing data of the second dataset.}
\label{tab:A-LSTM_2_comparison}
\resizebox{\textwidth}{!}{%
\begin{tabular}{llccc}
\toprule
\textbf{Our Signal Preprocessing} & \textbf{Model} & \textbf{RMSE $\downarrow$} & \textbf{\( \bm{R}^2 \)} $\uparrow$ & \textbf{MAPE (\%) $\downarrow$} \\
\midrule
\multirow{4}{*}{W/o $\dot{Q}$} 
& A-LSTM & 174.83 & 0.84 & 8.14 \\
& A-LSTM + CNN & 167.37 & 0.85 & 7.62 \\
& A-LSTM + CNN + ODE-A-LSTM & 168.51 & 0.84 & 8.01 \\
& A-LSTM + CNN + LSTM & 166.50 & 0.85 & 7.70 \\
& A-LSTM + CNN + ODE-LSTM (Ours) & 163,31 & 0,85 & 7,54 \\
\midrule
\multirow{4}{*}{With $\dot{Q}$} 
& A-LSTM & 168.09 & 0.87 & 7.47 \\
& A-LSTM + CNN & 169.02 & 0.87 & 7.40 \\
& A-LSTM + CNN + ODE-A-LSTM & 170.69 & 0.87 & 8.03 \\
& A-LSTM + CNN + LSTM & 165.00 & 0.87 & 7.30 \\
& A-LSTM + CNN + ODE-LSTM (Ours) & \textbf{158.09} & \textbf{0.89} & \textbf{6.95} \\
\bottomrule
\end{tabular}
}
\end{table*}

Figure~\ref{fig:rmse_combined_delta} (left) compares several reparameterizations:
\(\dot{Q}_i(I,Q)=\operatorname{Interp1D}(I_{i,j},Q_{i,j})\big|_{I_k}\),
\(\dot{V}_i(V,Q)=\operatorname{Interp1D}(V_{i,j},Q_{i,j})\big|_{V_k}\),
\(\dot{I}_i(I,Q)=\operatorname{Interp1D}(I_{i,j},Q_{i,j})\big|_{I_k}\),
\(\dot{Q}_i(Q,V)=\operatorname{Interp1D}(Q_{i,j},V_{i,j})\big|_{Q_k}\),
\(\dot{V}_i(V,I)=\operatorname{Interp1D}(V_{i,j},I_{i,j})\big|_{V_k}\), and
\(\dot{I}_i(I,V)=\operatorname{Interp1D}(I_{i,j},V_{i,j})\big|_{I_k}\).
Our proposed feature is \(\dot{Q}_i(I,Q)\). Note that all these derived features are applied during the signal preprocessing stage, up to and including the delta-feature computation. The results show that using only derived features yields the highest RMSE (252.52), indicating that relying solely on derived features leads to poor predictive performance.
In contrast, the feature combination \([I,V,Q,\dot{Q}(I,Q)]\) achieves the lowest RMSE (162.37), outperforming all other configurations.
The strong contribution of \(\dot{Q}\) underscores the effectiveness of Capacity Derivative Tracking for modeling discharge data: by aligning capacity values with corresponding current measurements, it provides a more consistent representation of the current–capacity relationship across cycles and improves predictive accuracy.

In the delta feature computation step, the difference in extracted features between each cycle and the one occurring \textit{delta} steps earlier is computed. The choice of the \textit{delta} parameter is important, as it determines the temporal resolution of feature variation. To assess its impact, we evaluated \textit{delta} values ranging from 1 to 10, as shown in Figure~\ref{fig:rmse_combined_delta} on the right. Among these, a \textit{delta} value of 9 achieved the best performance, yielding the lowest RMSE of 162.37.

\subsection{Validation of Prediction Model}\label{subsec:prediction}
As shown in Table~\ref{tab:A-LSTM_2_comparison}, the comparison between using only statistical features ($I$, $V$, $Q$) without $\dot{Q}$ and using combined fused features ($I$, $V$, $Q$, $\dot{Q}$) demonstrates that the proposed signal preprocessing pipeline—enhanced with the derived capacity feature $\dot{Q}$—consistently improves performance across all evaluated models. Notably, our prediction model, combining CNN, A-LSTM, and ODE-LSTM, achieves the best performance with the lowest RMSE of 158.09, highest \( R^2 \) score of 0.89, and lowest MAPE of 6.95\%. These results confirm the effectiveness of integrating tailored CNN and ODE-LSTM blocks with A-LSTM, particularly when enriched by the derived capacity feature. Moreover, the results indicate that incorporating ODE into A-LSTM (ODE-A-LSTM) does not yield superior performance compared to the use of standard LSTM in this context.

\begin{figure*}[!t]
  \centering
  \includegraphics[width=1\textwidth]{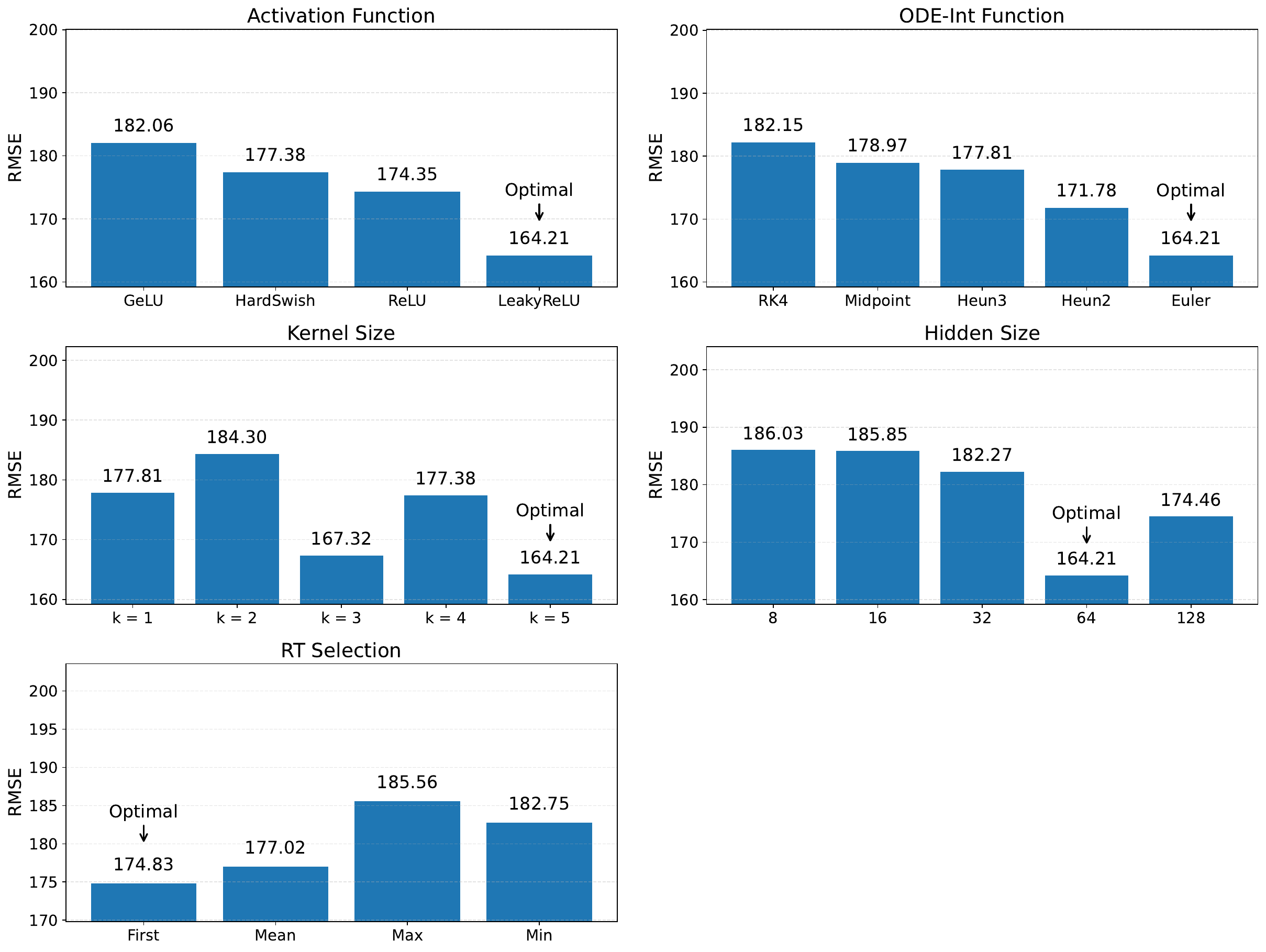}
  \caption{RMSE Comparison Across Configurations of the Proposed Prediction Model.}
  \label{fig:rmse_model_configs}
\end{figure*}

To assess the impact of various configurations on the prediction model performance, we conducted a series of experiments evaluating RMSE across different settings. These included activation functions (GeLU~\cite{lee2023mathematical}, HardSwish~\cite{avenash2019semantic}, ReLU~\cite{banerjee2019empirical}, LeakyReLU), ODE integration schemes (RK4~\cite{tay2012spreadsheet}, Midpoint~\cite{yu2019properties}, Heun2/3~\cite{saadoon2025numerical}, Euler~\cite{salleh2012ordinary}), convolutional kernel sizes, hidden layer sizes, and the selection of the relative time variable $R_t$ (RT), as defined in the A-LSTM paper~\cite{liu2020anomaly}. Each configuration was isolated while keeping other components fixed to ensure fair comparison. The results, visualized in Figure~\ref{fig:rmse_model_configs}, indicate that specific choices—such as using LeakyReLU as the activation function, Euler for ODE integration, a kernel size of $k=5$, a hidden size of 64, and first RT in A-LSTM block yielded lower RMSE values. The optimal settings consistently yield the lowest RMSE, indicating that these configurations better capture battery degradation signatures and, in turn, improve RUL prediction accuracy.

\begin{figure*}[!t]
  \centering
  \includegraphics[width=1\textwidth]{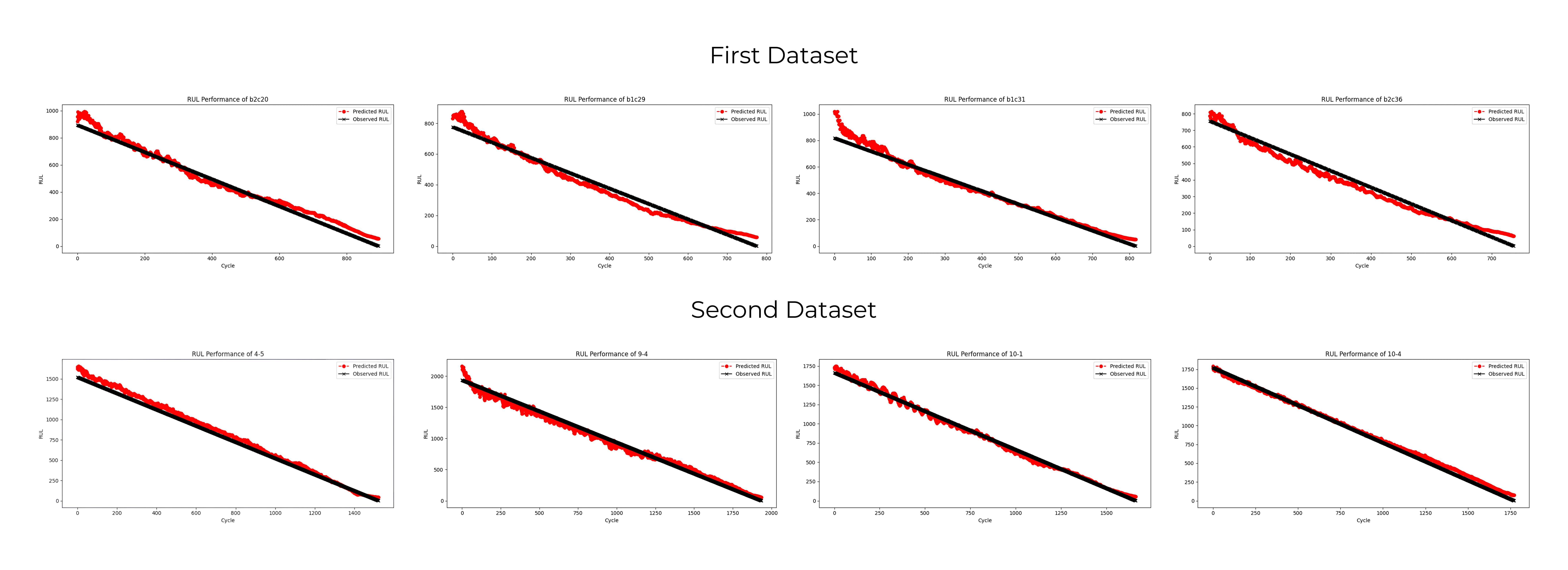}
  \caption{Comparison of predicted RUL and actual RUL for some batteries from the first and the second datasets.}
  \label{fig:visualization_RUL}
\end{figure*}

Figure~\ref{fig:visualization_RUL} illustrates the RUL prediction performance of the proposed model across eight test battery cells taken from the test sets of two battery datasets. The red curves represent the predicted RUL, while the black curves denote the ground truth. Across all battery cells, the predicted RUL closely follows the actual degradation trend, confirming the model’s ability to generalize across diverse data distributions. Notably, the model successfully captures the nonlinear degradation behavior, particularly in cells such as \texttt{b1c31} and \texttt{10-4}, where the actual RUL exhibits mild fluctuations. Although slight overestimations appear in the early cycles and near EOL periods, the predictions remain consistently aligned with observed values. 

Table~\ref{tab:existing_methods_comparison} compares the proposed hybrid framework with several baselines:  traditional machine learning models (ElasticNet~\cite{hans2011elastic}, XGBoostRegression~\cite{wang2022xgboost}), Transformer-based deep learning model~\cite{vaswani2017attention}, and current state-of-the-art (SOTA) approaches~\cite{ma2022real, ge2024structural}, across two battery datasets. For the traditional machine learning models, the input data has been processed using our proposed signal preprocessing pipeline. On the first dataset, the proposed model significantly outperforms other models ElasticNet, XGBoostRegression, and the Transformer, achieving the lowest RMSE of 101.59, the highest \( R^2 \) score of 0.82, and a substantially reduced MAPE of 8.29\%. Similarly, on the second dataset, the proposed model surpasses all other methods, including the state-of-the-art baselines from \cite{ma2022real, ge2024structural}, with an RMSE of 158.09, an \( R^2 \) score of 0.89, and the lowest MAPE of 6.95\%. These results highlight the effectiveness and generalizability of the proposed approach across different data distributions. Compared with the two SOTA approaches~\cite{ma2022real, ge2024structural}, which use a simple machine learning model and a shallow CNN that is insufficient to capture complex battery degradation dynamics, our prediction model adopts a hybrid architecture: a CNN + A-LSTM branch for characteristic feature extraction, an LSTM branch to capture temporal order, and an ODE-based component to model continuous degradation behavior. Although the resulting architecture is more expressive, the signal preprocessing pipeline reduces the small input size, keeping the overall model complexity compatible with practical hardware constraints (as detailed in Table~\ref{tab:efficiency}).

\begin{table*}[!t]
\centering
\caption{Performance comparison between the proposed approach, traditional machine learning models, and the state-of-the-art methods \cite{ge2024structural, ma2022real} on the second dataset.}

\label{tab:existing_methods_comparison}
\resizebox{\textwidth}{!}{%
\begin{tabular}{lllccc}
\toprule
\textbf{Dataset} & \textbf{Signal Preprocessing} & \textbf{Prediction Model} & \textbf{RMSE $\downarrow$} & \textbf{\( \bm{R}^2 \)} $\uparrow$ & \textbf{MAPE (\%) $\downarrow$} \\
\midrule

\multirow{5}{*}{First dataset} 
& Ours &  ElasticNet  & 319.98  &  -0.20 & 26.70\\ 
& Ours & Transformers  & 193,83 & 0,47 & 16,59\\
& Ours & XGBoostRegression  &  156.33 & 0.70 & 12.19\\
& Ours & Ours & \textbf{101.59} & \textbf{0.82} & \textbf{8.29} \\

\midrule

\multirow{5}{*}{Second dataset} 
& Ours &  ElasticNet &  575.54 & -0.20 & 26.67\\
& Ours & Transformers  & 196,75 & 0,79 & 9,16 \\
& Ours & XGBoostRegression  & 169.88 & 0.87 & 7.37 \\
& \cite{ma2022real} & \cite{ma2022real}  &  186 & 0.804 & 8.72 \\
& \cite{ge2024structural} & \cite{ge2024structural}  & 192.17 & 0.858 & -\\
& Ours & Ours  & \textbf{158.09} & \textbf{0.89} & \textbf{6.95} \\

\bottomrule
\end{tabular}
}
\end{table*}

\begin{figure}[t]
    \centering
    \begin{subfigure}{0.48\textwidth}
        \centering
        \includegraphics[width=\linewidth]{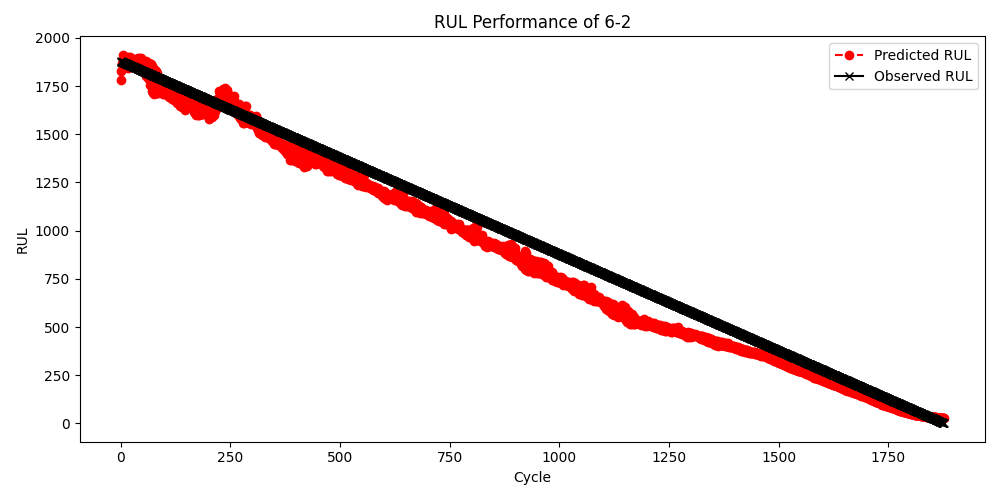}
        \caption{With ODE-LSTM}
        \label{fig:ODE_LSTM_6_2}
    \end{subfigure}
    \hfill
    \begin{subfigure}{0.48\textwidth}
        \centering
        \includegraphics[width=\linewidth]{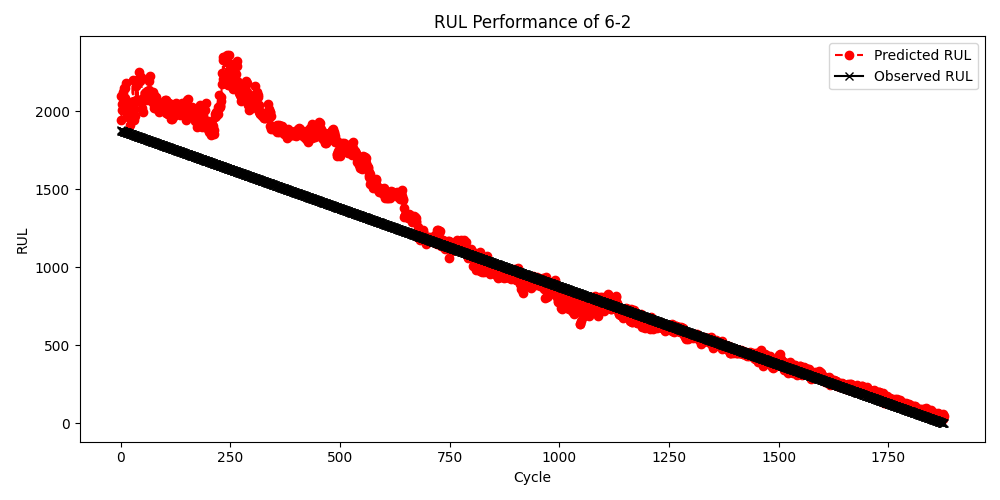}
        \caption{With LSTM}
        \label{fig:LSTM_6_2}
    \end{subfigure}
    \caption{Comparison of our prediction model using ODE-LSTM and LSTM.}
    \label{fig:with_ODE_LSTM_comparison}
\end{figure}

Figure~\ref{fig:with_ODE_LSTM_comparison} compares the proposed ODE-LSTM with a standard LSTM.  
Unlike the standard LSTM, the ODE-LSTM introduces a continuous-time hidden state transition, enabling the latent dynamics to evolve smoothly between discrete observations. This results in more stable predictions across consecutive cycles, avoiding the abrupt fluctuations observed with the standard LSTM.

\begin{figure}[t]
    \centering
    \begin{subfigure}{0.48\textwidth}
        \centering
        \includegraphics[width=\linewidth]{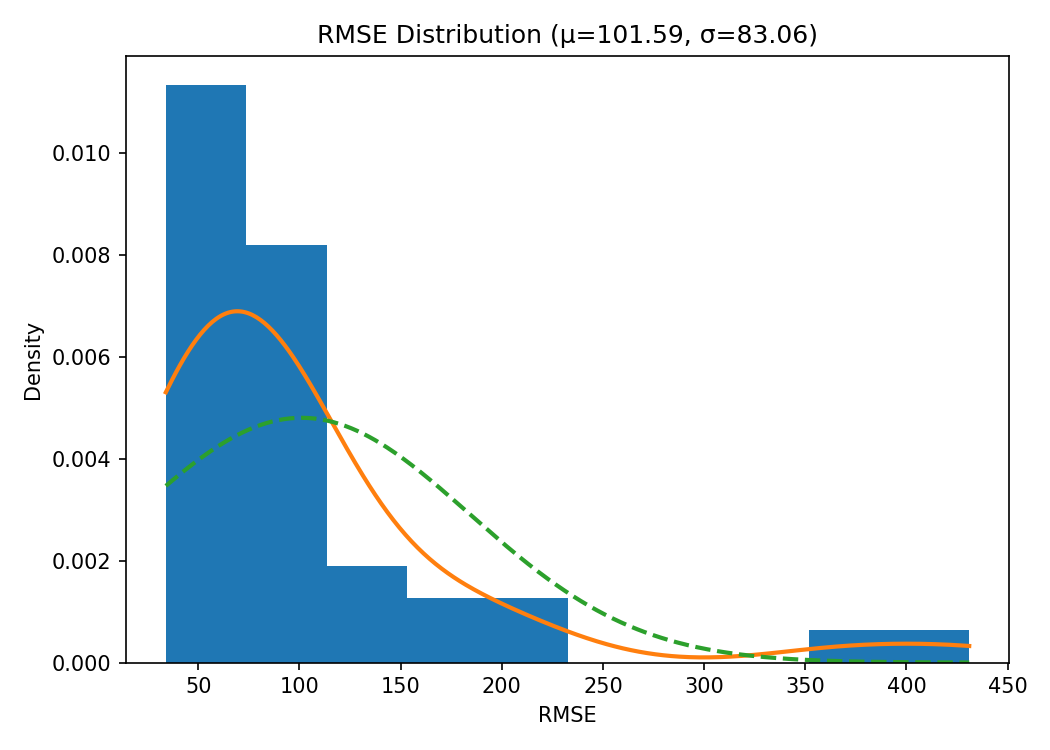}
        \caption{First Dataset}
        \label{fig:First_rmse_hist_kde}
    \end{subfigure}
    \hfill
    \begin{subfigure}{0.48\textwidth}
        \centering
        \includegraphics[width=\linewidth]{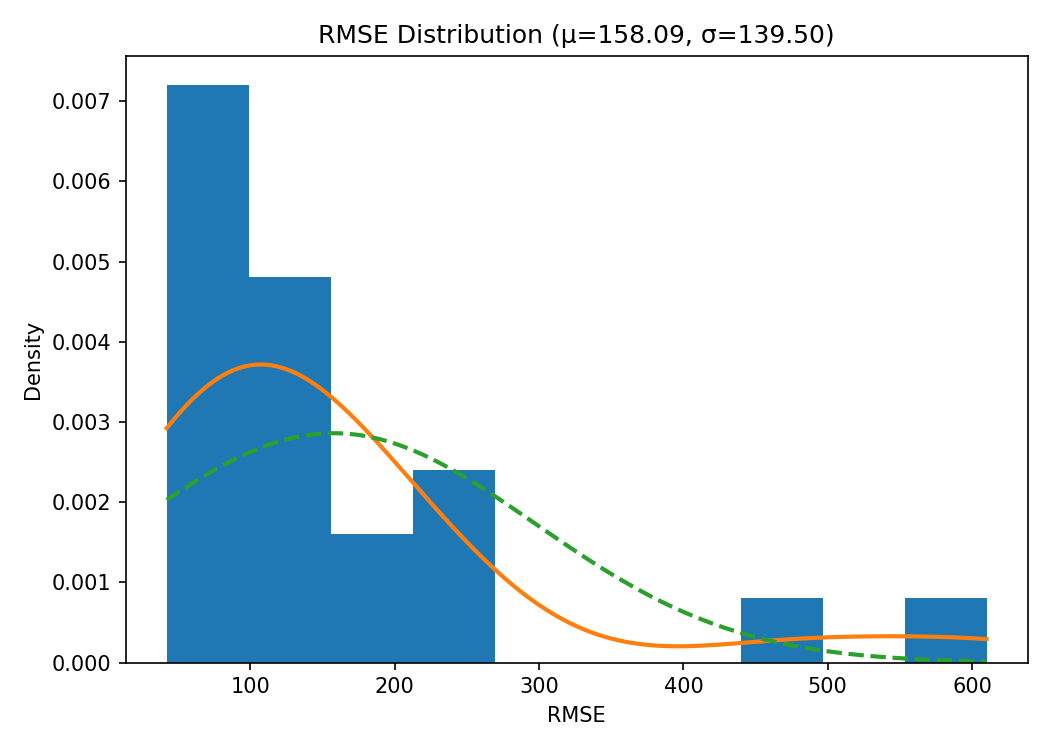}
        \caption{Second Dataset}
        \label{fig:Second_rmse_hist_kde}
    \end{subfigure}
    \caption{Across-cell RMSE distributions on the test sets}
    \label{fig:rmse_hist_kde}
\end{figure}

Figure \ref{fig:rmse_hist_kde} shows the RMSE distributions across cells on the testing data of the first and the second dataset. Bars indicate the empirical density, the solid curve is a Kernel Density Estimate (KDE)~\cite{chen2017tutorial} and the dashed curve is a fitted normal. The errors concentrate at low–to–moderate RMSE with a right-skewed tail driven by a few difficult cases. On the first dataset, the mean RMSE is 158.09 and most cells fall below RMSE=200, indicating accurate and stable RUL prediction with occasional outliers. On the second dataset, the mean RMSE decreases to 101.59, and the narrower histogram indicates improved consistency across cells.

The results in Table~\ref{tab:efficiency} demonstrate that the proposed model is computationally efficient and suitable for real-world deployment. With only 1.53M parameters and a compact model size of 5.84~MB, the architecture has a small memory footprint that can be deployed on resource-constrained devices. Furthermore, on an AMD Ryzen~5 CPU, the measured inference latency for processing a single sample is below 4~ms (p90), with a throughput of nearly 287 inferences per second on a single core. This indicates that the model can deliver real-time predictions without requiring specialized hardware. Overall, these results validate that the proposed method achieves both high accuracy and practical efficiency, supporting its applicability for real-world RUL prediction tasks.

\begin{table}[t]
\centering
\caption{Computational efficiency of the proposed model.}
\begin{tabular}{lcc}
\toprule
\textbf{Metric} & \textbf{Value} \\
\midrule
Parameters & 1,531,521 \\
Model size (FP32) & 5.84 MB \\
CPU latency (p50) & 3.49 ms \\
CPU latency (p90) & 4.12 ms \\
Mean latency & 3.64 ms \\
Throughput & 286.9 inferences/s \\
\bottomrule
\end{tabular}
\label{tab:efficiency}
\end{table}

\subsection{Extended Validation of the Prediction Model for Early Cycle Life Prediction}
\begin{figure}[t]
    \centering
    \includegraphics[width=0.5\textwidth]{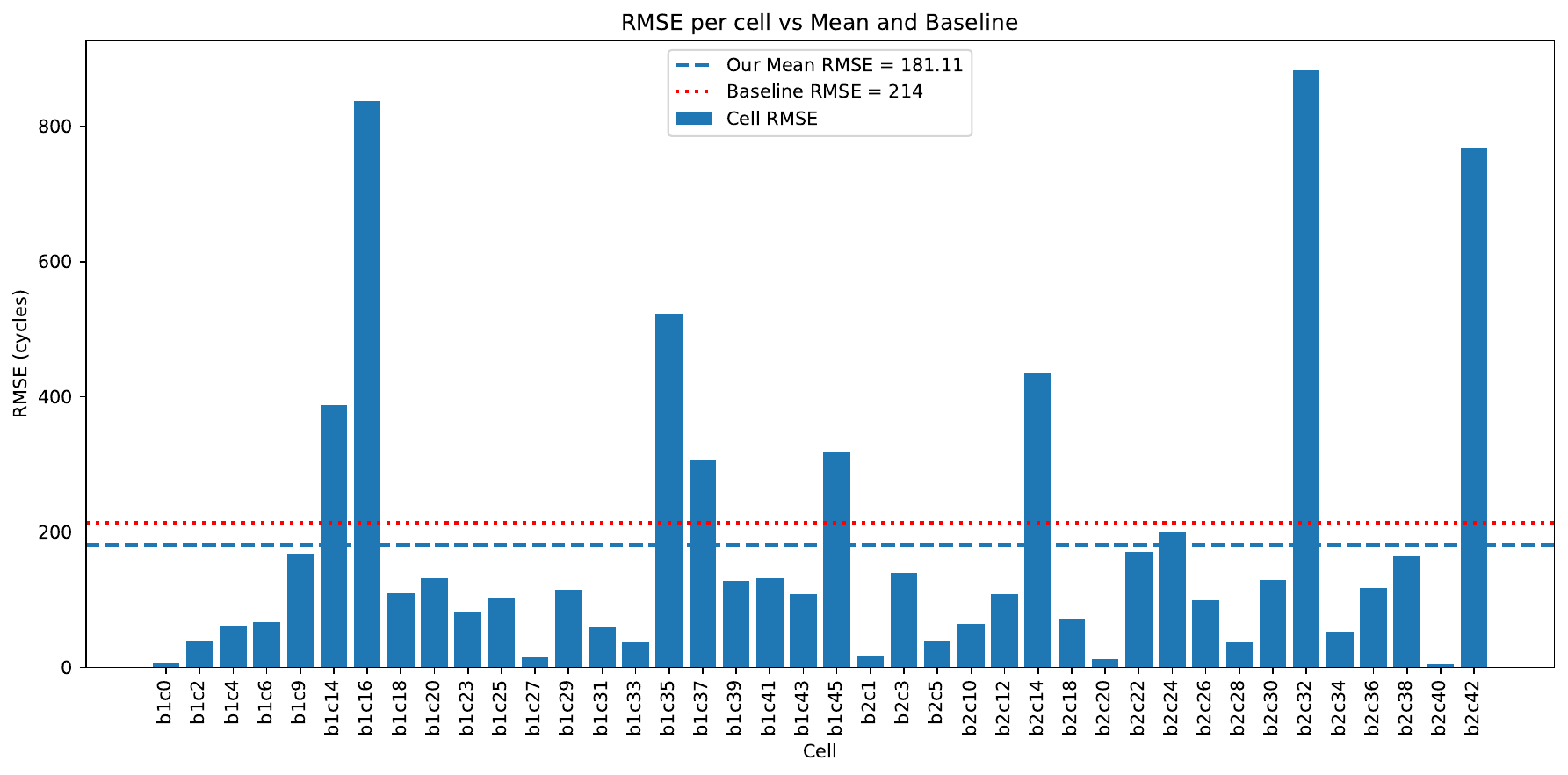}%
    \caption{RMSE of cycle life early prediction (in cycles).}

    \label{fig:RMSE_cycle_life}
\end{figure}
Figure~\ref{fig:RMSE_cycle_life} shows the RMSE of early cycle-life prediction on the first dataset. The bars represent the per-cell RMSE, the dashed line indicates our mean RMSE across all cells (181.11 cycles), and the dotted line marks the performance of the baseline model~\cite{severson2019data} (the original paper for the first dataset) at 214 cycles (as reported in Table~1 of~\cite{severson2019data}). Although we use the same experimental setup as~\cite{severson2019data}, including the same data splitting, their model uses the first 100 cycles as input, whereas our model uses only the first 30 cycles. Despite this, our model still outperforms the baseline, demonstrating its strong overall performance for both RUL estimation and early cycle-life prediction.

\subsection{Extended Validation of Transfer Learning on the Prediction Model}

To evaluate the generalization capability of the proposed model across different battery datasets, we conducted a series of transfer learning experiments as detailed in Table~\ref{tab:TF_description}. In cases involving upper and lower half splits, the training data were divided into two groups of battery cells based on cycle life: cells with a cycle life below the median of the cycle lives of the training set were assigned to the lower half, while those above the median were placed in the upper half. These experiments explore various scenarios in which the model is pre-trained on one dataset and fine-tuned on another, simulating real-world applications where labeled data in the target domain is scarce.

\begin{table*}[!t]
\centering
\caption{Configuration of transfer learning strategies for each case, showing the datasets used for pre-training and fine-tuning with their respective testing sets.}
\label{tab:TF_description}
\resizebox{\textwidth}{!}{%
\begin{tabular}{llll}
\toprule
\textbf{Testing data of} & \textbf{Case} & \textbf{Pre-Training} & \textbf{Fine-Tuning} \\
\midrule

\multirow{5}{*}{First dataset} 
& 1   & Entire training data of the first dataset     & -- \\
& 2   & Entire training data of the second dataset    & -- \\
& 3.1 & Entire training data of the second dataset    & Upper half of training data from the first dataset \\
& 3.2 & Entire training data of the second dataset    & Lower half of training data from the first dataset \\
& 4   & Entire training data of the second dataset    & Entire training data of the first dataset \\

\midrule

\multirow{5}{*}{Second dataset} 
& 5   & Entire training data of the second dataset    & -- \\
& 6   & Entire training data of the first dataset     & -- \\
& 7.1 & Entire training data of the first dataset     & Upper half of training data from the second dataset \\
& 7.2 & Entire training data of the first dataset     & Lower half of training data from the second dataset \\
& 8   & Entire training data of the first dataset     & Entire training data of the second dataset \\

\bottomrule
\end{tabular}
}
\end{table*}

First, we use Case 8 to investigate the impact of different block-freezing strategies during transfer learning, shown in Table~\ref{tab:TL_method_comparison}. Notably, freezing only the A-LSTM block yields the best overall performance, suggesting that this component captures more transferable temporal features. In contrast, freezing all three blocks (CNN, A-LSTM, and ODE-LSTM) results in severe performance degradation, underscoring the importance of model flexibility during fine-tuning. Based on these findings, we adopt this block-freezing strategy of freezing only A-LSTM block for the next experiment, as shown in Table~\ref{tab:TF_performance_comparison}.

\begin{table}[!t]
\centering
\caption{Performance comparison under various block-freezing strategies applied during transfer learning in Case 8 (see Table~\ref{tab:TF_description}).}
\label{tab:TL_method_comparison}
\begin{tabular}{lcccc}
\toprule
\textbf{Transfer Learning Strategy} & \textbf{RMSE $\downarrow$} & \textbf{\( \bm{R}^2 \)} $\uparrow$ & \textbf{MAPE (\%) $\downarrow$} \\ 
\midrule
Freezing CNN block (1) & 166.18 & 0.88 & 7.21 \\
Freezing A-LSTM block (2) & \textbf{157.92} & \textbf{0.88} & \textbf{6.88} \\
Freezing ODE-LSTM block (3) & 159.64 & 0.87 & 6.79 \\
(1) + (2) & 170.05 & 0.86 & 7.48 \\
(1) + (3) & 179.4 & 0.85 & 8.04 \\
(2) + (3) & 161.98 & 0.86 & 7.01 \\
(1) + (2) + (3)  & 282.69 & 0.69 & 12.79 \\
\bottomrule
\end{tabular}
\end{table}

The results, summarized in Table~\ref{tab:TF_performance_comparison}, demonstrate that training on one dataset and evaluating on the other without transfer learning (Cases 2 and 6) yields markedly poor performance, with high RMSE and negative \( R^2 \), indicating a significant mismatch in data distributions between the two datasets.

However, introducing fine-tuning with partial data from the target domain (Cases 3.1, 3.2, 7.1, and 7.2) noticeably improves model accuracy, with Case 7.2 achieving a strong balance (RMSE = 201.63, \( R^2 = 0.81 \), MAPE = 8.63\%). Full fine-tuning on the target domain after pre-training (Cases 4 and 8) yields performance close to or even better than training directly on the target dataset (Cases 1 and 5), with Case 8 achieving the best results on the second dataset (RMSE = 157.92, \( R^2 = 0.88 \), MAPE = 6.88\%).

\begin{table}[!t]
\centering
\caption{Performance comparison of our proposed model on different training strategies with Transfer Learning on both first and second dataset.}
\label{tab:TF_performance_comparison}
\begin{tabular}{cccc}
\toprule
\textbf{Case} & \textbf{RMSE $\downarrow$} & \textbf{\( \bm{R}^2 \) $\uparrow$} & \textbf{MAPE (\%) $\downarrow$} \\ 
\midrule

1 & \textbf{101.59} & \textbf{0.82} & \textbf{8.29} \\  
2 & 543.56 & -2.58 & 46.66 \\
3.1 & 160.98 & 0.61 & 14.4 \\
3.2 & 290.75 & 0.00 & 23.31 \\
4 & 123.44 & 0.76 & 10.64 \\ 

\midrule

5 & 158.09 & \textbf{0.89} & 6.95 \\
6 & 800.23 & -1.21 & 35.76 \\
7.1 & 231.12 & 0.73 & 11.69 \\
7.2 & 201.63 & 0.81 & 8.63 \\
8 & \textbf{157.92} & 0.88 & \textbf{6.88} \\

\bottomrule
\end{tabular}
\end{table}

\subsection{Discussion}
In this study, we focused on evaluating the effectiveness of the proposed model under multiple charge and discharge profiles, while the batteries in the laboratory LFP/graphite lithium-ion datasets were tested under controlled, constant ambient temperature conditions. The results demonstrate that the model performs well across various charging conditions in the first dataset and across various discharging conditions in the second dataset. In future work, we plan to extend this study by (i) incorporating both cell and ambient temperature variations using real-world EV driving datasets, and (ii) testing additional battery datasets with different positive electrode materials, such as nickel–manganese–cobalt oxides (NMC), to further assess the chemical generalization capability of the model.

Furthermore, the current model utilizes only voltage and current signals as inputs. In subsequent research, we aim to integrate additional features derived from graph-based representations~\cite{zhao2025new, he2025dcaggcn}. For instance, in an EV battery pack, each sensor corresponding to a battery cell can be defined as a node, enabling the construction of a graph or hypergraph that captures inter-cell coupling effects (e.g., temperature correlations, shared cooling paths). These graph-structured features will then be combined with the proposed prediction model to enhance spatial–temporal degradation modeling.

Moreover, the proposed model has been evaluated using transfer learning between two datasets: the first containing various charge profiles and the second containing various discharge profiles. In future work, we will (i) extend this evaluation to cross-operating-condition transfer using real-world EV data (e.g., driving scenarios with acceleration, braking, and regenerative events), (ii) investigate the impact of different end-of-life thresholds (e.g., 70\% vs.\ 80\% capacity) on prediction accuracy, and (iii) explore uncertainty-aware or heteroscedastic loss functions to better characterize prediction confidence, particularly in late high-risk stages.

Finally, the predicted battery RUL provides actionable insights for optimizing charging and discharging strategies, scheduling battery replacement or recycling, and ultimately extending battery lifespan while reducing user anxiety in electric vehicle applications. In future work, we also plan to study how the proposed model can be embedded into on-board battery management systems with realistic computational and memory constraints.

\section{Conclusion}\label{sec:conclusion}
This study presents an robust approach for predicting the RUL of lithium-ion batteries by integrating an advanced signal processing pipeline with a novel prediction model. The signal preprocessing pipeline, comprising Capacity Interpolation and Denoising, Statistical Feature Extraction, Delta Feature Computation, and Feature Fusion, effectively captures the degradation dynamics over charge-discharge cycles. The proposed prediction model, combining CNN, A-LSTM, and ODE-LSTM blocks, demonstrates strong capabilities in modeling both discrete and continuous time behaviors of battery aging. Notably, the integration of our tailored CNN and ODE-LSTM block enhances the prediction model's ability. The prediction model is further evaluated using transfer learning across different learning strategies and source data partitioning scenarios. Results indicate that the model maintains robust performance even when trained on limited data, underscoring its adaptability and effectiveness in data-constrained environments. Experimental validation on two large-scale datasets confirms that our method significantly outperforms baseline deep learning approaches and traditional machine learning techniques, indicating good adaptability in data-constrained settings. Future work will port the prediction model to ExecuTorch for on-device inference on edge hardware, enabling practical deployment in predictive maintenance systems. We will also evaluate the method’s generalizability across different battery chemistries and usage scenarios, and pursue real-time edge deployment.

\section*{Preprint Availability}
A preprint of this manuscript is available at:
\url{https://arxiv.org/pdf/2505.16664}

\section*{Contact Information}
For access to the code and further information about this proposed system, please contact AIWARE Limited Company at: \url{https://aiware.website/Contact}

\bibliographystyle{plain}
\bibliography{cas-refs}

\begin{IEEEbiography}[{\includegraphics[width=1in,height=1.25in,clip,keepaspectratio]{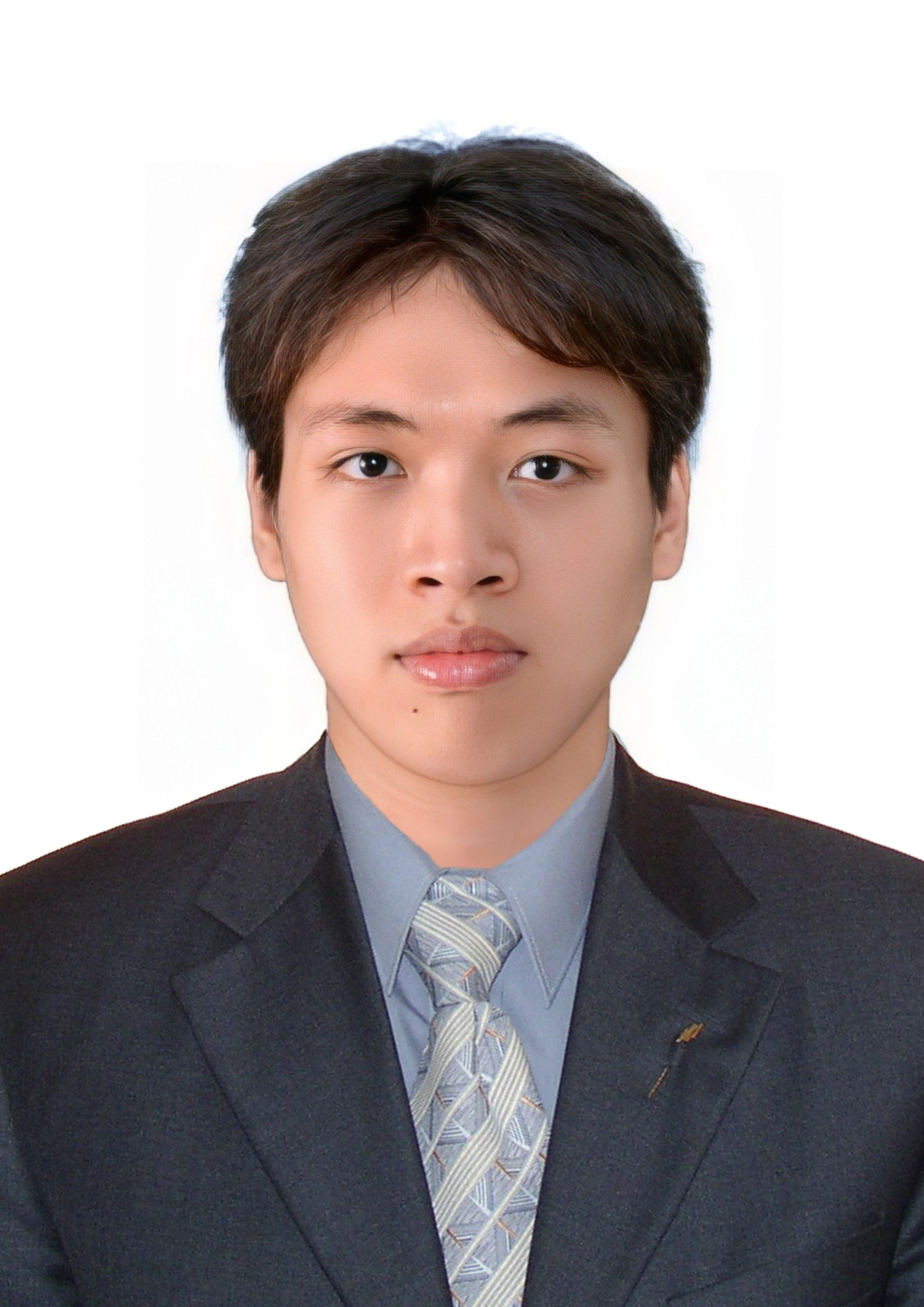}}]{KHOA TRAN} received the Engineering degree in Control Engineering and Automation from the University of Science and Technology – The University of Danang, Vietnam, in 2024. His current research interests include machine learning applications in battery management systems and predictive maintenance for rotating machinery.
\end{IEEEbiography}

\begin{IEEEbiography}[{\includegraphics[width=1in,height=1.25in,clip,keepaspectratio]{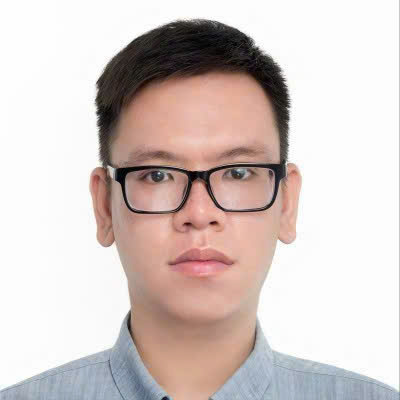}}]{TRI LE} received the degree of Engineer in Control Engineering and Automation from the University of Science and Technology, the University of Da Nang, in 2023. He is currently working as an AI/ML engineer. His research interests include battery and machinery aging and maintenance.
\end{IEEEbiography}

\begin{IEEEbiography}[{\includegraphics[width=1in,height=1.25in,clip,keepaspectratio]{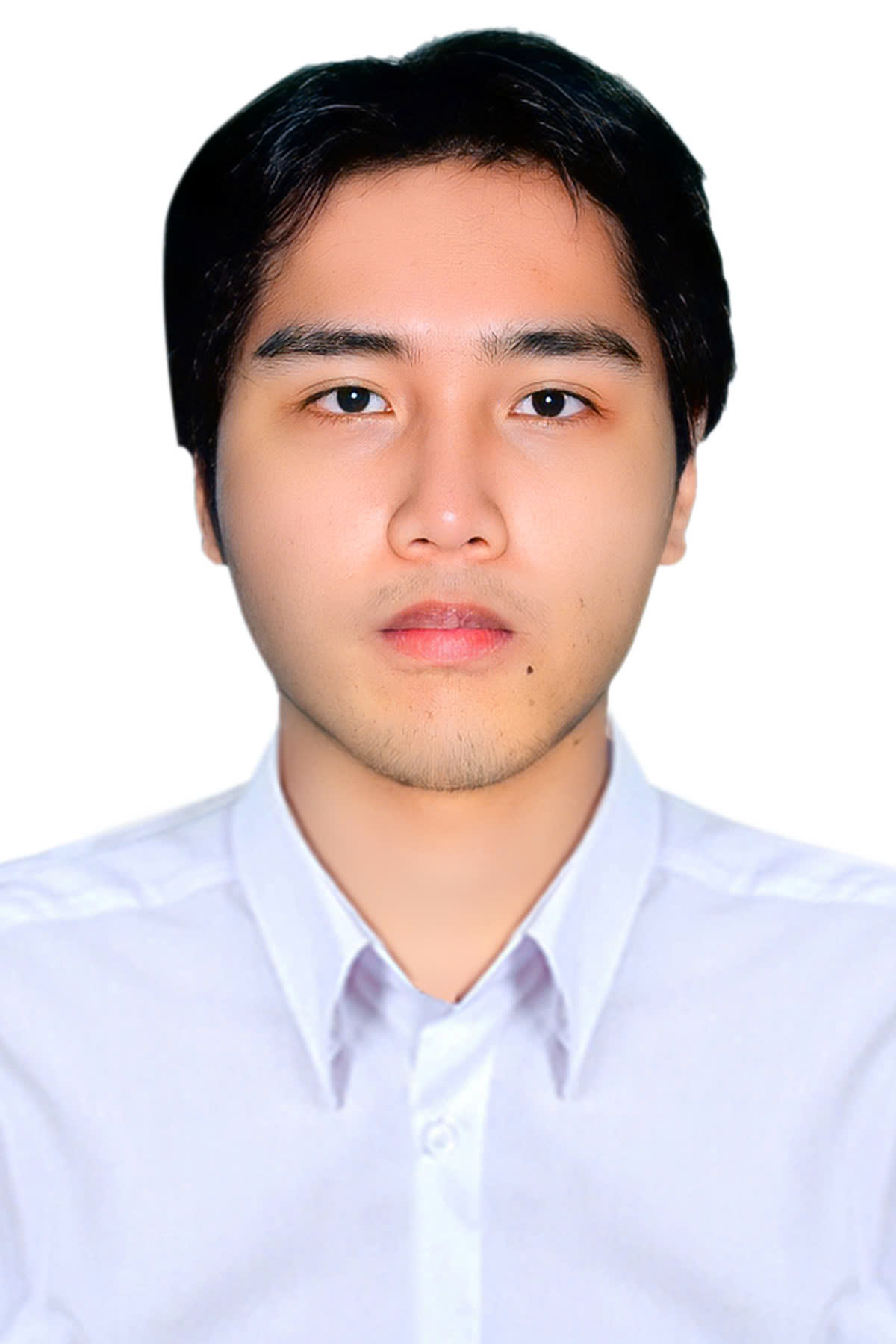}}]{BAO HUYNH} received a engineer's degree in Data Science and Artificial Intelligence from The University of Da Nang - University of Science and Technology (DUT) in 2025. Currently, he is pursuing a Master’s in Computer Science at DUT with a research focus. His research interests include deep learning applications in image processing, image reconstruction, and predicting the lifespan of engines or batteries.
\end{IEEEbiography}

\begin{IEEEbiography}[{\includegraphics[width=1in,height=1.25in,clip,keepaspectratio]{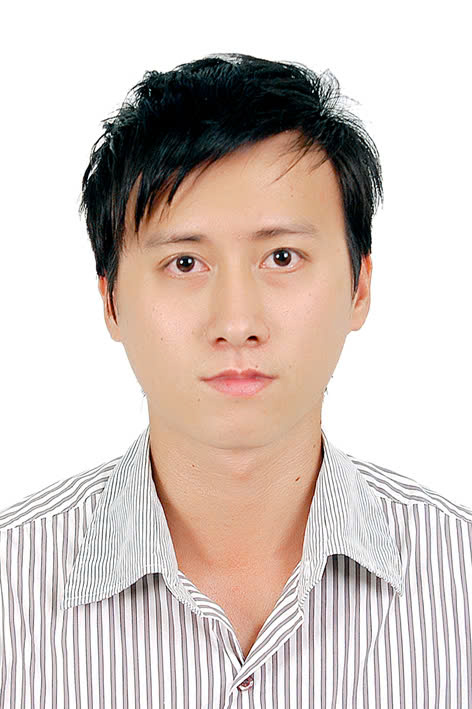}}]{HUNG CUONG-TRINH} received the Ph.D. degree in computer engineering from the
University of Ulsan, South Korea, in 2017. Since 2018, he has been a Lecturer and a
Researcher in the Faculty of Information Technology, Ton Duc Thang University, Ho Chi
Minh City. His research interests include bioinformatics, artificial intelligence, data science, and machine learning applications.
\end{IEEEbiography}

\begin{IEEEbiography}[{\includegraphics[width=1in,height=1.25in,clip,keepaspectratio]{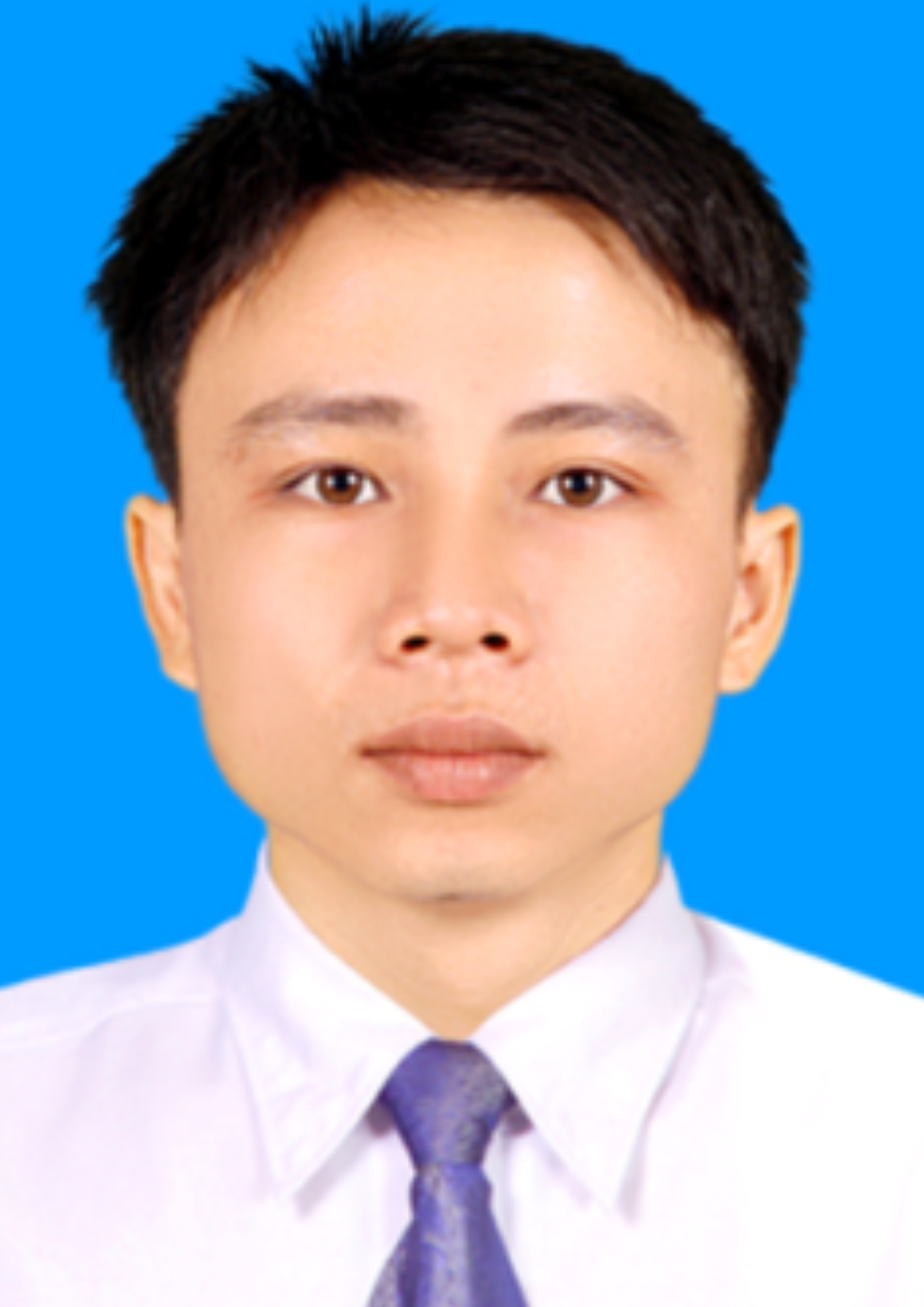}}]{VY-RIN NGUYEN} received his B.S. and M.S. degrees in Computer Science. He has taught and conducted research in Computer Science, Embedded Systems, Microchip Design, and Software Engineering at institutions such as FPT University, the Advanced Institute of Science and Technology (AIST – The University of Da Nang), and Duy Tan University. He has led and participated in multiple research projects funded by the Ministry of Education and Training, as well as at the provincial and university levels.

His research interests include analog IC design (energy-efficient microchips), FPGA-based systems (parallel processing and multithreading), embedded systems, and broadband communication. He is currently collaborating with international research centers in the USA, Germany, Japan, and South Korea, and with industry partners such as AMD (Xilinx), TSMC, Cadence, and Synopsys.
\end{IEEEbiography}

\begin{IEEEbiography}[{\includegraphics[width=1in,height=1.25in,clip,keepaspectratio]{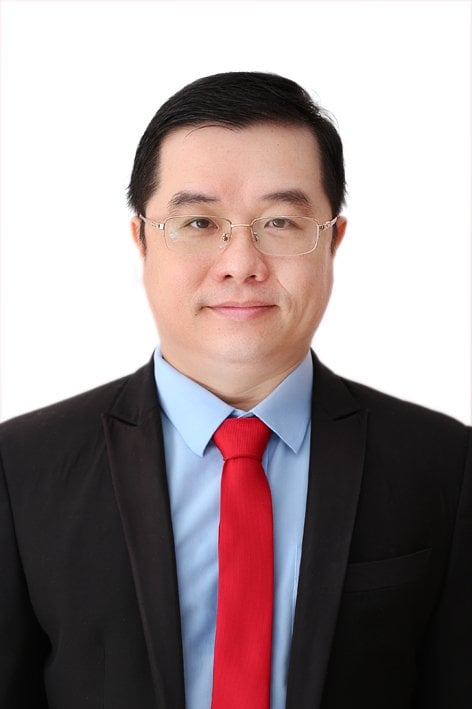}}]{T. NGUYEN-THOI}
received the Ph.D. degree in Mechanical Engineering from the National University of Singapore in 2010, where he was awarded the Best Ph.D. Thesis. He is currently an Associate Professor and the Director of the Institute for Computational Science \& Artificial Intelligence at Van Lang University (VLU), Vietnam.

His research interests include Computational Mechanics, Applied Mathematics, and Artificial Intelligence. He has published over 350 ISI-indexed papers and completed six national research projects. He has an H-index of 73 with more than 19,000 citations according to ISI.

Dr. Nguyen-Thoi is a member of the Scientific Council of the National Foundation for Science and Technology Development (NAFOSTED), specializing in Mechanics – Engineering for the terms 2020–2022, 2022–2024, and 2025–2027. He was awarded the “Bualuang ASEAN Chair Professorship” (2022–2024) by the Faculty of Engineering, Thammasat University, Thailand.

He has been recognized in the top 1\% of Highly Cited Researchers (Clarivate Analytics, Web of Science, USA) in 2021 and has been listed among the top 2\% of most-cited scientists worldwide for six consecutive years (2019–2024) by Stanford University. He is also the recipient of the “Lifetime Achievement Award” from Ton Duc Thang University in 2017.

Dr. Nguyen-Thoi serves on the Editorial Board of the journal Computers \& Structures (Elsevier, Q1) and several other international journals.
\end{IEEEbiography}

\begin{IEEEbiography}[{\includegraphics[width=1in,height=1.25in,clip,keepaspectratio]{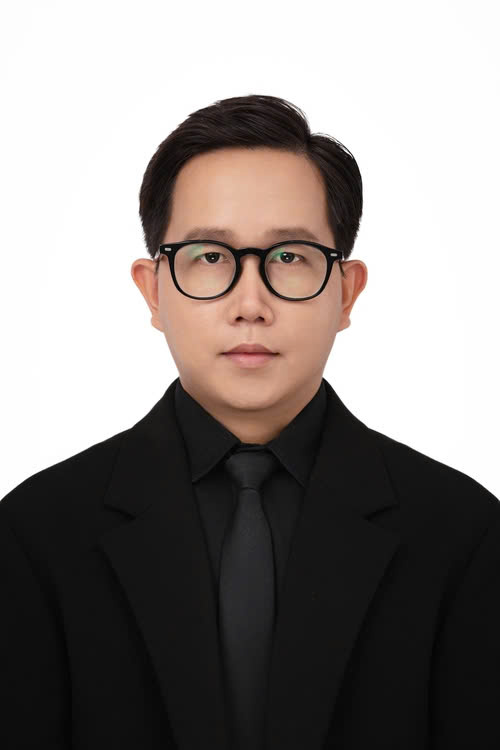}}]{VIN NGUYEN-THAI}
received his Bachelor of Civil Engineering from Ton Duc Thang University, Vietnam in 2018 and his Master of Civil Engineering from the same university in 2021. Since July 2024, he has been serving as a Research Assistant at the Institute for Computational Science \& Artificial Intelligence, Van Lang University. His research interests include Structural Damping, Structural Optimization, Computational Mechanics, Finite Element Analysis, and Artificial Intelligence.
\end{IEEEbiography}

\EOD

\end{document}